\documentclass[lettersize,journal]{IEEEtran}
\usepackage{amsmath,amsfonts}
\usepackage{algorithmic}
\usepackage{algorithm}
\usepackage{array}
\usepackage[caption=false,font=normalsize,labelfont=sf,textfont=sf]{subfig}
\usepackage{textcomp}
\usepackage{stfloats}
\usepackage{url}
\usepackage{verbatim}
\usepackage{graphicx}
\usepackage{cite}
\usepackage{multirow}
\usepackage{color}
\definecolor{citeco}{RGB}{0,174,239}
\usepackage[colorlinks,
linkcolor=red,       
anchorcolor=blue,  
citecolor= citeco,        
]{hyperref}
\begin{document}

\title{Salient Object Detection From Arbitrary Modalities}

\author{Nianchang Huang, Yang Yang, Ruida Xi, Qiang Zhang*, Jungong Han, Jin Huang
\thanks{Nianchang Huang, Yang Yang and Qiang Zhang are with the State Key Laboratory of Electromechanical Integrated Manufacturing of High-Performance Electronic Equipments, and the Center for Complex Systems, School of Mechano-Electronic Engineering, Xidian University, Xi’an, Shaanxi 710071, China. Email: huangnianchang@xidian.edu.cn, yang@stu.xidian.edu.cn, ruidaxi@stu.xidian.edu.cn, and qzhang@xidian.edu.cn. }
\thanks{Jungong Han is with the Department of Computer Science, University of Sheffield, U.K. Email: jungonghan77@gmail.com }
\thanks{Jin Huang is with the State Key Laboratory of Electromechanical Integrated Manufacturing of High-Performance Electronic Equipment, Xi’an, Shaanxi 710071, China. Email: jhuang@mail.xidian.edu.cn.}
\thanks{*Corresponding authors: Qiang Zhang.} }	

\markboth{Journal of \LaTeX\ Class Files,~Vol.~14, No.~8, August~2021}%
{Shell \MakeLowercase{\textit{et al.}}: A Sample Article Using IEEEtran.cls for IEEE Journals}


\maketitle

\begin{abstract}
Toward desirable saliency prediction, the types and numbers of inputs for a salient object detection (SOD) algorithm may dynamically change in many real-life applications. 
However, existing SOD algorithms are mainly designed or trained for one particular type of inputs, failing to be generalized to other types of inputs.  
Consequentially, more types of SOD algorithms need to be prepared in advance for handling different types of inputs, raising huge hardware and research costs.
Differently, in this paper, we propose a new type of SOD task, termed Arbitrary Modality SOD (AM SOD). 
The most prominent characteristics of AM SOD are that the modality types and modality numbers will be arbitrary or dynamically changed.
The former means that the inputs to the AM SOD algorithm may be arbitrary modalities such as RGB, depths, or even any combination of them. 
While, the latter indicates that the inputs may have arbitrary modality numbers as the input type is changed, \emph{e.g.} single-modality RGB image, dual-modality RGB-Depth (RGB-D) images or triple-modality RGB-Depth-Thermal (RGB-D-T) images. 
Accordingly, a preliminary solution to the above challenges, \emph{i.e.} a modality switch network (MSN), is proposed in this paper.  
In particular, a modality switch feature extractor (MSFE) is first designed to extract discriminative features from each modality effectively by introducing some modality indicators, which will generate some weights for modality switching. 
Subsequently, a dynamic fusion module (DFM) is proposed to adaptively fuse features from a variable number of modalities based on a novel Transformer structure.
Finally, a new dataset, named AM-XD, is constructed to facilitate research on AM SOD.
Extensive experiments demonstrate that our AM SOD method can effectively cope with changes in the type and number of input modalities for robust salient object detection.

\end{abstract}


\section{Introduction}\label{sec::Intro}

\textcolor{blue}{\IEEEPARstart{S}{alient} object detection (SOD) aims at detecting the most visually attractive objects from the inputs \cite{10130326, BAO2023126560}, which has been widely performed on many computer vision tasks, such as tracking \cite{apptrack}, segmentation \cite{appseg1, appseg2}, action recognition \cite{appact}, camouflaged object detection \cite{FPNF}, and so on.}

\begin{figure}[!t]
	\centering
	\includegraphics[width=\linewidth]{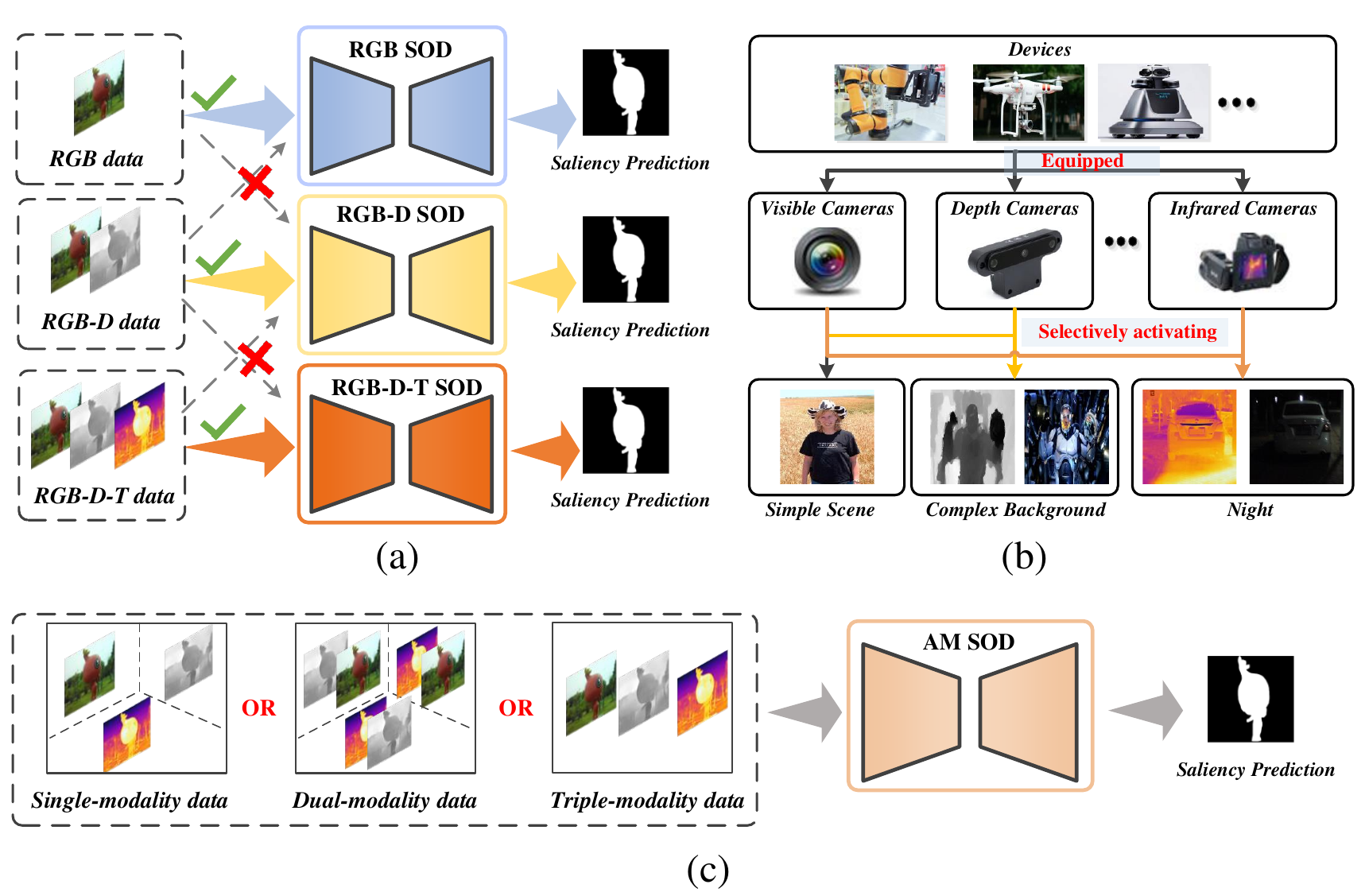}
	\caption{RGB SOD, RGB-D/T SOD vs AM SOD. (a) Existing SOD models. (b) Devices with multiple cameras. (c) AM SOD. The modality types and numbers for the inputs of existing SOD models must be fixed, while the modality types and numbers for our proposed AM SOD model may be arbitrary or changed.
	}
	\label{fig_AMSOD}
\end{figure}

Recently, how to exploit the complementary information within multi-modal images for breaking the bottleneck of RGB SOD has attracted extensive attention. Specifically, RGB images can capture rich 2D spatial information from the scenes, but drop abundant 3D spatial information. Meanwhile, they cannot capture scene information well under inadequate lighting conditions. 
Consequentially, most existing RGB SOD models may work well for those simple scenes but for complex scenarios.   
Over the last decade, different imaging sensors, such as depth sensors and thermal sensors, have been rapidly developed and widely applied in various devices. Depth images or thermal images can complement RGB images by providing more scene information, \emph{e.g.} 3D spatial information within depth (D) images or shape information within thermal (T) images, which is beneficial for  conquering some limitations of RGB SOD algorithms mentioned above.
As a result of that, multi-modal (\emph{e.g., } RGB-D or RGB-T) SOD algorithms have been developed rapidly and have already achieved significant progress in recent year\cite{k01, k02,r42,n50,n51}.

Although multi-modal SOD model have greater advantages than RGB SOD ones, they may also have their own limitations, especially when applied to some real-life applications. 
Specifically, as shown in Fig. \ref{fig_AMSOD}(a), existing multi-modal SOD models are specifically designed or trained for some particular modalities, \emph{e.g.} RGB-D SOD models for RGB-D images or RGB-T SOD models for RGB-T images. \textcolor{blue}{In this paper, we term such SOD task as modality-specific (MS) SOD. }
When their inputs' modalities are changed, \emph{e.g.} feeding RGB-T images into RGB-D SOD models, those models will not work well. This cannot satisfy the demand on dynamically changing modality types for some special real-life applications. Specifically, as shown in Fig. \ref{fig_AMSOD}(b), many devices, such as robots, smartphones, and drones, may be equipped with multiple cameras of different modalities in practical applications.
While the users will not activate all cameras for SOD in every scenario, and only selectively activate some of the cameras according to the specific conditions for pursuing better performance or lower costs. 
For example, they may only use RGB cameras to detect salient objects for those simple scenarios, employ thermal cameras for some night time scenes, or activate RGB cameras and depth cameras simultaneously for complex scenes. 
Considering that, the users may have to prepare multiple SOD algorithms to handle different types of inputs. This will inevitably introduce more hardware and research costs.

A natural question is that: \emph{is it possible to detect salient objects from the images with arbitrary modality types or arbitrary modality numbers by using one algorithm?} If so, those real-life applications can use one model to process different types of inputs without introducing many additional costs. 
Toward this goal, we consider a new taks, i.e., Arbitrary Modality salient object detection (AM SOD) in this paper, which aims at taking the images of arbitrary modality types and arbitrary modality numbers as the inputs and output the corresponding salient objects by using just one model (Fig. \ref{fig_AMSOD}(c)). 


In particular, the significant characteristic of AM SOD over existing ones is that \textit{the inputs may be one arbitrary modality type}, rather than one specific type, from all of possible modality types, \emph{i.e.} one of $2^M$-1 modality types for $M$ modalities.
Besides, the derived characteristic from arbitrary modality types is \textit{arbitrary modality numbers}. That is, the inputs may have arbitrary modality numbers as the input modality type is changed, \emph{e.g.} one image for RGB inputs, two images for RGB-D inputs, and three images for RGB-D-T inputs.
Note that, such two characteristics are also two inherent challenges when deploying the AM SOD algorithm.
Taking such two challenges into account, in this paper, we tackle them in a holistic manner by proposing a modality switch network (MSN), where two key issues need to be addressed.

\textbf{Modality discrepancies in unimodal feature extraction, i.e., how to adaptively extract discriminative features from arbitrary modalities by using one feature extractor?}


Different from existing SOD models, which mainly focus on extracting features from one or two certain modalities, AM SOD generally has to handle a broader range of image modalities. 
Accordingly, AM SOD models need to learn more diverse feature distributions, since their imaging mechanisms of different modalities are essentially distinct.
One straightforward approach is to employ an independent feature extractor for each modality. 
However, this approach will significantly increase the model parameters as the types of image modalities increase, which is undesirable for those resource-limited devices.
Alternately, a compromised approach is to utilize a siamese networks (i.e., only one to-be-trained feature extractor) to extract different modal features simultaneously. 
This will obviously reduce the model parameters.
Nevertheless, the discrepancies among different modalities will prevent an AM SOD model from accurately fitting the distributions of all modalities simultaneously and tend to favor their common ones when just one modality-shared feature extractor is adopted, thus easily leading to suboptimal results.

As a trade-off between model complexity and representational ability for different modalities, a modality switch feature extractor (MSFE) will be presented in our proposed MSN to adaptively extract unimodal features from arbitrary modalities by only using one network. 
Specifically, a modality indicator will be first introduced for each modality in our MSFE to adaptively extract those unimodal features according to their modalities. Here, the modality indicators for the images of the same modality will be the same, and those for different modalities will be orthogonal to each other. Then, instead of using multiple feature extractors,  our proposed MSFE will generate some weights from those modality indicators to switch our feature extractor for capturing different types of unimodal features. Finally, a modality-aware loss will be specially designed to ensure such switch process by predicting the modalities of those extracted unimodal features. By doing so, our proposed MSFE will possess the ability to adaptively extract discriminative features from arbitrary modalities.

\textbf{Dynamic inputs in multi-modal feature fusion, i.e., how to dynamically fuse those unimodal features from an arbitrary number of modalities?
}

The core of multi-modal saliency object detection (SOD) lies in the effective integration of complementary information across modalities to enhance saliency prediction performance. 
Toward this goal, many elaborate multi-modal information fusion methods have been proposed in recent years, yielding encouraging results.
However, it should be noted that, different from the existing SOD models, the modality types and modality numbers of unimodal features are dynamically changed for AM SOD models, \emph{e.g.} two types of unimodal features for RGB-D/RGB-T images, and three types of unimodal features for RGB-D-T images.
However, existing fusion methods typically rely on specific modal types as well as fixed modal numbers, which are inadequate in coping with such dynamic input changes.
On the one hand, the design motivation of some existing fusion strategies necessitates that the inputs modality type should not be changed arbitrarily. 
For example, considering that leveraging the abundant geometric knowledge from depth maps is helpful to provide object localization for SOD, Li et al. \cite{n41} emphasize the use of depth information to enhance RGB information in their proposed Cross-Modal Weighting (CMW) module.
Therefore, the arbitrary alternation of input modality types may lack rationality for such specific fusion strategy.
On the other hand, the efficacy of some existing fusion methods may be compromised when the number of input modalities increases or decreases due to the fixed network structures or specific network weights.
Therefore, it is necessary to explore some new dynamic fusion paradigms for AM SOD.

Drawing inspiration from the popular Transformer structures, a novel dynamic fusion module (DFM) will be specially designed to address such an issue in this paper. Specifically, our proposed DFM will leverage Transformer's capability of processing an arbitrary number of input tokens to achieve dynamic fusion for an arbitrary number of modalities. For that, our proposed DFM will first treat the unimodal features of each modality as a distinct token. Accordingly, the number of modalities for the inputs will be equal to the number of tokens in our proposed DFM. Then, a novel cross-modal attention strategy will be designed in our proposed DFM, which will first explore channel-wise interactions across the features of modalities and subsequently fuse those unimodal features for effectively capturing their complementary information. By doing so, our proposed DFM can dynamically fuse those unimodal features from an arbitrary number of modalities for detecting salient objects. Besides, we conduct a new AM SOD dataset, \emph{i.e.} AM-XD, based on some existing SOD datasets to facilitate research on AM SOD. 

In summary, the main contributions of this work are as follows: 

(1) We propose a novel SOD task, termed Arbitrary Modality SOD, which aims at detecting salient objects from the input images of arbitrary modality types and modality numbers, \emph{e.g.} RGB images, RGB-D images, RGB-T images and RGB-D-T images. 

(2) We analyze the two major challenges, \emph{i.e.} modality discrepancies in unimodal feature extraction and dynamic inputs in multi-modal feature fusion, for our new AM SOD research, and conduct a new dataset for further investigation and comparisons.

(3) We propose a preliminary solution, \emph{i.e.} a modality switch network for our new task, which uses a modality switch feature extractor to adaptively extract unimodal features from different types of images and a dynamic fusion module to dynamically capture the complementary information within the features across the input images of arbitrary numbers and modalities.

\section{Related work} \label{sec::RW}

\subsection{RGB SOD}

RGB SOD has been popular for quite some time now and aims at detecting salient objects from a single RGB image\cite{10056257}. Conventional models mainly design some handcrafted features or employ some prior knowledge (\emph{e.g.} center/boundary prior) for SOD. 

However, those conventional models cannot well extract those high-level semantic features, thus usually leading to sub-optimal results. In contrast, deep convolutional neural networks (DCNN) can automatically extract discriminative low- and high-level features from massive data. Accordingly, DCNN-based SOD models become mainstream\cite{RGB1, RGB2, RGB3, RGB4, RGB5}. 
For example, Liu et al. \cite{RGB2} presented a PoolNet+, which explores the potential of pooling techniques on SOD. They first design a global guidance module (GGM) to guide the location information of potential salient objects at different feature levels on the bottom-up pathway. Then, they propose a  feature aggregation module (FAM) to fuse the coarse-level semantic information with the fine-level features in the top-down pathway.
Recently, some DCNN-based SOD models try to accurately recover the boundaries of those predicted salient objects by introducing some boundary information. For example, Yao et al. \cite{RGB1} proposed a boundary information progressive guidance network (BIPGNet), where multiple saliency detection units (SDUs) are employed to enable simultaneous salient region and boundary detections.

\subsection{RGB-D SOD}

Recently, RGB-D SOD has witnessed remarkable advancements and lots of works have been published \cite{CATNet}. According to their ways of exploiting cross-modal complementary information, most existing RGB-D SOD can be roughly categorized into three classes, \emph{i.e.} pixel-level fusion, feature-level fusion, and decision-level fusion.

Pixel-level fusion based models usually first concatenate the RGB images and depth images as four-channel images and then feed them into a single-stream SOD model to detect salient objects \cite{n24, n25, n68, in1}. For instance, Chen et al. \cite{n25} proposed to temporally cascade RGB and depth images for exploring those cross-modal complementary information in the 3D feature space by using a modified 3D CNN. 

Feature-level fusion based models \cite{n50, r42, n51, n20, fea1, fea2 ,fea3, fea4, CIRNet,z27} usually first extract the unimodal RGB features and unimodal depth features from the input RGB images and depth images, respectively, by using two independent sub-networks. Then, they fuse the extracted unimodal RGB and depth features by designing some fusion modules. Finally, those fused features will be fed into a decoder to detect those salient objects. For example, Zhang et al. \cite{n20} proposed a novel Bi-directional Transfer-and-Selection Modules (BTS), which establishes an interactive relationship between RGB features and depth features through a bidirectional structure, thereby achieving better fusion results. 

Decision-level fusion based models \cite{r27, n103} first detect two preliminary saliency maps from the input RGB image and depth image, respectively, by using two SOD models. Then, they fuse the two maps as the final one by using some fusion strategies.   For example, Wang et al. \cite{n103} proposed to employ reinforcement learning to generate the weights for fusing the two primary saliency maps.

\subsection{RGB-T SOD}
The research on RGB-T SOD started relatively late compared to the above two ones. Until 2019, the first deep learning-based RGB-T SOD model \cite{k01} emerged. In the last two years, RGB-T SOD has attracted significant research interest.
Generally speaking, most existing RGB-T SOD \cite{RGBTSOD1, RGBTSOD2, k03, k04, RGBTSOD3, RGBTSOD4,RGBTSOD5} models are based on feature-level fusion. Most existing models primarily focus on enhancing multi-modal feature fusion strategies and refining saliency strategies.  For example, Chen et al. \cite{k03} introduced a modality transfer fusion (MTF) module, which first reduces the semantic gap between single and multi-modal images, and then captures the cross-modal complementary information by leveraging some point-to-point structural similarity information. While Wang et al. \cite{k04} proposed a Cross-Scale Alternate Guiding Fusion (CSAGF) module in the prediction stage, which effectively explores and combines high-level features from various scales.

\subsection{Modality Unified SOD}
In recent years, the use of complementary information, i.e., depth or thermal information, has shown its benefits for salient object detection (SOD). However, the RGB-D or RGB-T SOD problems are mainly solved independently. With the development of SOD, some researchers tend to simultaneously achieve the two types of multi-modal SOD tasks (i.e., RGB-D and RGB-T SOD tasks) in a unified framework. For example, Gao et al. \cite{Un1, Un2} proposed the first unified end-to-end framework for both RGB-D and RGB-T SOD.
Although such modality unified frameworks show their robustness and generality on different modality combinations (e.g., RGB-D and RGB-T), they have to be retrained to meet the task requirements when the input modality combinations change. Differently, our proposed AM SOD model aims to take images of arbitrary modality types and arbitrary modality numbers as inputs and robustly output the corresponding salient objects without additional training costs.

\textcolor{blue}{Recently, Jia et al. \cite{jia2023one} presented an all-in-one salient object detection model, which can process RGB SOD tasks, RGB-T SOD tasks, and RGB-D SOD tasks by using one model.
Besides, in some other computer vision fields, modality unified frameworks have also aroused the interest of researchers and made impressive progress \cite{CMXNet, vscode}. For example, VSCode \cite{vscode} explores the relations between SOD tasks and camouflaged object detection (COD) tasks and builds a unified framework for four SOD tasks and two COD tasks, recently. However, those works mainly focus on the aspect of task unified paradigm in the output terminal, aiming at solving different tasks by using one framework. Different from them, we mainly focus on the aspect of modality unified paradigm for SOD task, which pays more attention to the input terminal.  Specifically, although they may seem to be similar, we do not try to design a unified paradigm for different SOD tasks, such as RGB SOD, RGB-T SOD and RGB-D SOD. Alternatively, we try to detect salient objects from arbitrary modalities, whose task is fixed but its inputs are optional or unified for different modalities. Accordingly, this may provide more perspectives for other visual tasks.
}

\begin{figure*}[!t]
\centering
\includegraphics[width=0.95\linewidth]{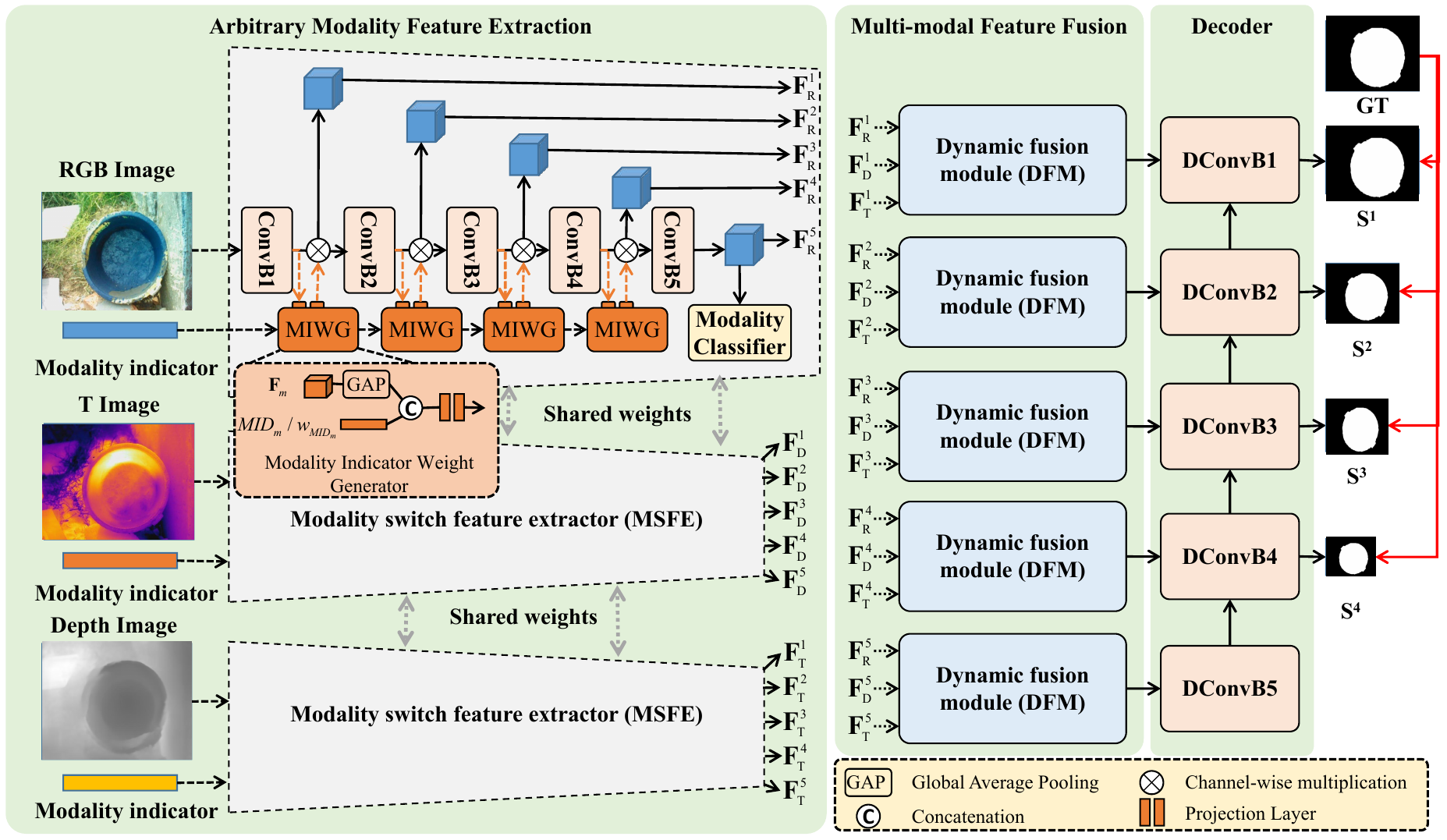}
\caption{Framework of our proposed MSN. The input images are first fed into the switch feature extractor to extract their unimodal features. Here, the number of input images is arbitrary. For better understanding, we display all the images of all modalities. However, the networks or features with dash lines are optional and may or may not exist. Then, the unimodal feature will be fused by using the dynamic fusion module. Finally, the saliency maps are obtained by using the saliency prediction decoder. }
\label{fig_Frame}
\end{figure*}

\section{Proposed Model}\label{sec::Model}

\subsection{Problem formulation}

Suppose that $\{M_1, M_2,..., M_K\}$ denotes $K$ modalities, such as RGB images, thermal images and depth images, event images, multispectral images, hyperspectral images, ultrasonic images, X-ray images, and so on. Accordingly, there are  $2^K-1$ types of modality combinations. Different from existing models which mainly focus on one specific type of inputs among the $2^K-1$ types, our proposed AM task aims to accurately detect salient objects from all of the above $2^K-1$ conditions (modality types) using just one model. In the following content, we will discuss the challenges within the fundamental process of AM SOD models.
 
\textbf{Feature Extraction:} The feature extractor of AM SOD models aims to extract discriminative unimodal features from different modalities, \emph{i.e.}
\begin{equation}
\label{eq_001}
\mathbf{F}_{m} = f_{extractor} ( \mathbf{X}_{m} ; \alpha_{extractor} ),
\end{equation}
where $\mathbf{F}_{m}$ denotes the unimodal features of $m$ modality. $f_{extractor} ( * ; \alpha_{extractor} )$ denotes the feature extractor with its parameters $\alpha_{extractor} $. $m\in \{M_1, M_2,..., M_K\}$. Here, the feature extractor $f_{extractor} ( * ; \alpha_{extractor} )$  should have the following properties. First, the feature extractor $f_{extractor} ( * ; \alpha_{extractor} )$ should be able to well handle the more diverse discrepancies among multiple modalities since AM SOD task may process more modalities than MS SOD models. Secondly, its network structure and parameters should not be  significantly changed when reducing or increasing modality numbers. For example, the straight way handling such modality discrepancies is to design an subnetwork for each modality. However, the structure of such feature extractor will be changed, \emph{e.g.} the two-stream network becomes the three-stream network when increasing $K$ from 2 to 3. Meanwhile, its parameters will be also significantly increased.

\textbf{Multimodel Fusion:} The fusion module aims at effectively capturing cross-modal complementary information across modaliies. Taking 3 modalities ($K=3$) as example, this process can be expressed by
\begin{equation}
\label{eq_002}
\mathbf{F}_{fuse} = \left\{\begin{array}{l}
g_{fuse} ( \mathbf{F}_{m_1}; \alpha_{fuse} ), m_1 \in  \{M_1, M_2, M_3\}, \\
g_{fuse} ( \mathbf{F}_{m_1},\mathbf{F}_{m_2} ; \alpha_{fuse} ), (m_1, m_2) \in  \{ (M_1, \\ M_2), (M_1, M_3) ,(M_3, M_2)\}, \\
g_{fuse} ( \mathbf{F}_{m_1}, \mathbf{F}_{m_2},\mathbf{F}_{m_3}; \alpha_{fuse} ), (m_1, m_2, m_3) \in  \\  \{(M_1, M_2, M_3)\},  \\
\end{array}\right.
\end{equation}
where $\mathbf{F}_{fuse}$ denotes the fused features. $g_{fuse} ( *; \alpha_{fuse} )$ denotes the fusion strategy with its parameters $\alpha_{fuse} $, which should have the following properties. (a) $g_{fuse} ( *; \alpha_{fuse} )$ can process a variable number of input features. (b) The fused features from  $g_{fuse} ( \mathbf{F}_{m_1}; \alpha_{fuse} )$, $g_{fuse} ( \mathbf{F}_{m_1},\mathbf{F}_{m_2} ; \alpha_{fuse} )$ and $g_{fuse} ( \mathbf{F}_{m_1}, \mathbf{F}_{m_2},\mathbf{F}_{m_3}; \alpha_{fuse} )$ should have the same dimensions/sizes, including width, height and channels. (c) $g_{fuse} ( *; \alpha_{fuse} )$ should satisfy the commutative law, i.e., $g_{fuse} ( \mathbf{F}_{m_1},\mathbf{F}_{m_2} ; \alpha_{fuse} )$ = $g_{fuse} ( \mathbf{F}_{m_2},\mathbf{F}_{m_1} ; \alpha_{fuse} )$ and $g_{fuse} ( \mathbf{F}_{m_1}, \mathbf{F}_{m_2},\mathbf{F}_{m_3}; \alpha_{fuse} )$ = $g_{fuse} ( \mathbf{F}_{m_1}, \mathbf{F}_{m_3},\mathbf{F}_{m_2}; \alpha_{fuse} ) =$,...,$=g_{fuse} ( \mathbf{F}_{m_3}, \mathbf{F}_{m_2},\mathbf{F}_{m_1}; \alpha_{fuse} )$. (d) The value range of the fused features for  $g_{fuse} ( \mathbf{F}_{m_1}; \alpha_{fuse} )$, $g_{fuse} ( \mathbf{F}_{m_1},\mathbf{F}_{m_2} ; \alpha_{fuse} )$ and $g_{fuse} ( \mathbf{F}_{m_1}, \mathbf{F}_{m_2},\mathbf{F}_{m_3}; \alpha_{fuse} )$  should also be the same.

\textbf{Saliency Prediction:} The prediction module aims to detect those salient objects from the fused features, \emph{i.e.}
\begin{equation}
\label{eq_03}
\mathbf{S} = f_{predictor} (\mathbf{F}_{fuse} ; \alpha_{prediction} ),
\end{equation}
where $f_{predictor} (*; \alpha_{prediction} )$ denotes the prediction module with its parameters $\alpha_{prediction}$. $\mathbf{S}$ denotes the saliency maps.

\subsection{Overall Framework}

As depicted in Fig. \ref{fig_Frame}, a preliminary solution, termed modality switch network (MSN), to the above AM SOD task is provided in this paper, which mainly consists of a modality switch feature extractor (MSFE), a dynamic fusion module (DFM) and a saliency prediction decoder.

Specifically, our proposed MSN takes as the inputs at least one image with arbitrary modality and its corresponding modality indicator. MSFE will first extract five levels of unimodal features from each input image. Then, our proposed DFM will automatically explore the relations among different types of unimodal features, and accordingly fuse them to capture their cross-modal complementary information. Finally, those fused features will be fed into the saliency prediction decoder to predict saliency maps. 
In this paper, we suppose that there are a total of three modalities, \emph{i.e.} RGB  modality, depth modality and thermal modality. The users can extend our proposed model to the data with more modalities without any extra modifications.  
We will detail each proposed module in the following content.

\subsection{Modality switch feature extractor (MSFE)}

As discussed in Section \ref{sec::Intro}, one of the major challenges for  AM SOD is how to effectively extract discriminative unimodal features from arbitrary modality by using one feature extractor, when facing modality discrepancies among multiple modalities.  This section proposes a modality switch feature extractor (MSFE) to address such an issue. 
As shown in Fig. \ref{fig_Frame}, besides the input images of different modalities, a modality indicator is introduced for each image to adaptively switch our proposed MSFE for extracting those discriminative unimodal features. \textcolor{blue}{ Moreover, we will unify one-dimension depth and thermal images into the three-dimension ones by using the color space conversion functions within the image processing library before feeding them into MSFE.}



Specifically, the inputs of our proposed MSFE are \{$\mathbf{X}_m^p$, $MID_m$\}$_{p=1}^{P}$, where $m \in \{R, D, T\}$ denotes RGB, depth or thermal modality, respectively. $\mathbf{X}_m^p$ denotes the $p$-th image of the modality m, and $MID_m$ denotes its corresponding modality indicator. It should be noted that the modality indicators $MID_m$ are the same for the images of the same modality, but are different for the images of different modalities. Here, the modality indicator $MID_R$ for RGB images is [1,0,0], $MID_D$ for depth images is [0,1,0], and $MID_T$ for thermal images is [0,0,1]. 
The procedures for extracting the first level of unimodal features are described as follows. 

First, the input image $\mathbf{X}_m$(we omit the index $p$ when it is clear) is fed into the first convolutional block to extract the unimodal features $\mathbf{\hat{F}}_m^1$, \emph{i.e.}
\begin{equation}
\label{eq_1}
\mathbf{\hat{F}}_m^1 = \operatorname{ConvB}(\mathbf{X}_m; \gamma_1),
\end{equation}
where $\operatorname{ConvB}(*; \gamma_1)$ denotes the convolutional block with its parameters $\gamma_1$. 

Then, the modality indicator $MID_m$ is fed into a Modality Indicator Weight Generator (MIWG) to generate some modality indicator weights $w_{MID_m}^1$ by
\begin{equation}
\label{eq_2}
w_{MID_m}^1 = \operatorname{Proj}(\operatorname{Cat}(MID_m, \operatorname{GAP}(\mathbf{\hat{F}}_m^1)); \theta_1),
\end{equation}
where $\operatorname{Proj}(*; \theta_1)$ denotes the projection layer with its parameters $\theta_1$. $\operatorname{Cat}(*)$ denotes concatenation operation. $\operatorname{GAP}(*)$ denotes global average pooling.
In this paper, the projection layer is conducted by using a fully connected layer stacked with a $\operatorname{Sigmoid}$ function. Besides, it should be noted that the parameters $\gamma_1$, $\theta_1$ and other parameters that will be introduced in the following contents are shared for all modalities, rather than independent for each modality as in existing models. 

Finally, the modality indicator weights $w_{MID_m}^1$ are used to enforce our proposed model to focus on those  characteristics according to the input modality, thus obtaining the first level of unimodal features $\mathbf{F}_m^1$, \emph{i.e.}
\begin{equation}
\mathbf{F}_m^1 =w_{MID_m}^1 \otimes  \mathbf{\hat{F}}_m^1,
\end{equation}
where $\otimes$ denotes the channel-wise multiplication.
From Eq. (5), it can be seen that the modality indicator weights $w_{MID_m}^1$ serve as a kind of modality switch for our proposed MSN by indicating a peculiar set of features according to the input modality. By doing so, our proposed MSN can adaptively extract those discriminative unimodal features from arbitrary modalities.
 
Similarly, the first level of unimodal features $\mathbf{F}_m^1$ and the first level of modality indicator weights $w_{MID_m}^1$ are fed into the next convolutional block and the next projection layer, respectively, to obtain the second level of $\mathbf{F}_m^2$ and $w_{MID_m}^2$ in the same way. Accordingly, the unimodal features and their modality indicator weights of the other three levels can also be obtained.
It should be noted that, in this paper, the feature extractor is based on the pre-trained ResNet50\cite{ResNet50}. Specifically, the convolutional blocks from 1 to 5 follow the same structures as the first 5 convolutional blocks of the ResNet50\cite{ResNet50} network. 

Moreover, in order to ensure those generated modality indicator weights be identifiable, an extra  modality-aware loss is further employed in MSN. For that, an extra classifier is first performed on the 5-th level of unimodal features to predict their modalities, \emph{i.e.}
\begin{equation}
MS_m = \operatorname{Softmax}(\operatorname{Linear}(\operatorname{GAP}(\mathbf{F}_m^5); \delta )),
\end{equation}
where $MS_m$ denotes the predicted probability that the input image belongs to the modality m.  $\operatorname{Softmax}(*)$ denotes the $\operatorname{Softmax}$ function. $\operatorname{Linear}(*; \delta )$ denotes a linear layer with its parameters $\delta $ for classification. $\operatorname{GAP}(*)$ denotes the global average pooling. 
Then,  a cross-entropy loss function is used to supervise  this classifier in the training stage by  
\begin{equation}
L_{modality} = \operatorname{CE}(MS_m, Y),
\end{equation}
where $ \operatorname{CE}(*)$ denotes the cross-entropy loss and $Y$ denotes the real modality of inputs. 

By doing so, those modality indicator weights will tell the feature extractor the specific modality of input images during feature extraction process. Accordingly, the feature extractor will learn to handle the inputs of different modalities, thus reducing modality conflicts. \textcolor{blue}{Moreover, the structure and parameters of our proposed MSN will not be significantly changed  when increasing or reducing the number of modalities.  }

\subsection{Dynamic fusion module (DFM) }
\begin{figure*}[!t]
\centering
\includegraphics[width=0.95\linewidth]{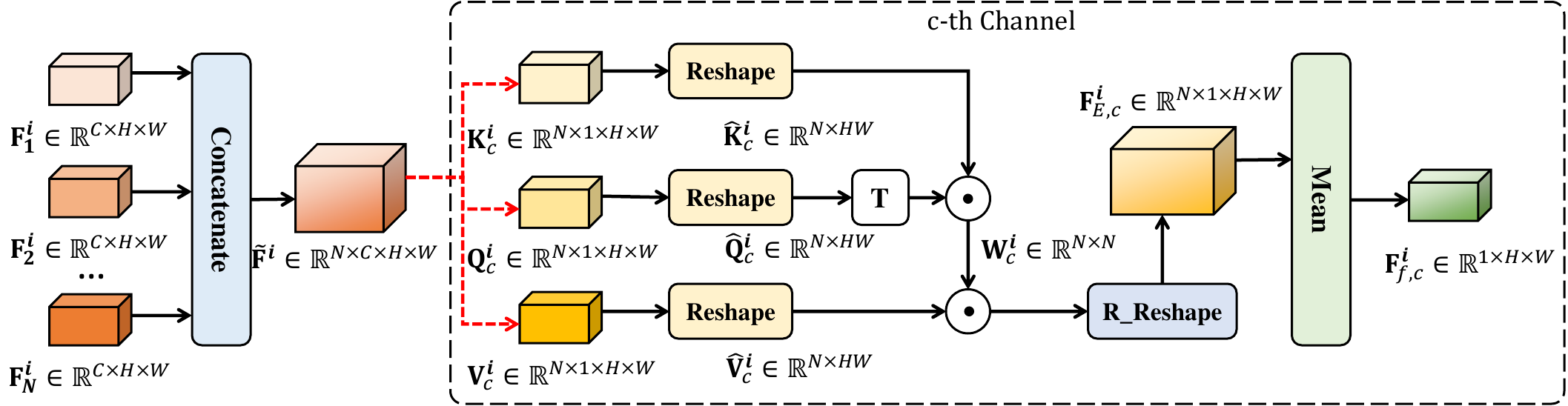}
\caption{Framework of our proposed DFM. }
\label{fig_IFM}
\end{figure*}

The primary goal of multi-modal SOD is to exploit multi-modal complementary information for breaking the limitations of unimodal images on salient object detection. 
However, as discussed in  Section I, existing multi-modal information fusion modules are inadequate in coping with their basic demands of AM SOD models due to the fixed network structures and specific network weights, which cannot handle inputs with an arbitrary number of modalities. 
As a result of that, existing fusion modules usually need to adjust their network structures according to the number of inputs and modality types, which will inevitably result in increased hardware and research costs.
For that, we propose a dynamic fusion module (DFM), as shown in  Fig. \ref{fig_IFM}, which can efficiently handle such variations in the number of input modalities and adaptively extract complementary information from any given multi-modal input images without changing its structure.
Specifically, it will explore the channel-wise relations across different unimodal features to capture their cross-modal complementary information within the multiple modalities effectively. More details are as follows.

 
\textcolor{blue}{Suppose that the inputs contain $N$ modalities. Here, $N$ should be less than the total number of assumed modalities in this paper, i.e., $K$ as mentioned above. Taking three modalities, \emph{e.g.} RGB, depth and thermal, as the example, the inputs are one-modality images (RGB images, depth images or thermal images) when $N=1$, two-modality images (RGB-D images, RGB-T images, and D-T images) when $N=2$ and three-modality images (RGB-D-T images) when $N=3$. }
The corresponding unimodal features, which are extracted from such input images, are collected and denoted as $\mathbf{F}_n^i \in \mathbb{R}^{C\times H \times W}$.
Here, $n=1,2,..,N$ denotes the features extracted from the image of the $n$-th modality.  $i=1,2,..,5$ denotes the $i$-th level of features. 
$C$ is the feature channel, representing different types of features. 
$H$ and $W$ are the height and width of features. 
Our proposed DFM will dynamically fuse them via the following steps.

Specifically, if the input is one image, \emph{i.e.} $N=1$, the fused or the output of DFM can be obtained by
\begin{equation}
	\mathbf{F}^{i}_{f}   =  \operatorname{Conv}(\mathbf{F}^i; \epsilon_v).
\end{equation}
Here, $\operatorname{Conv}(*; \epsilon_v)$ denotes a $1\times 1$ convolutional layers with its parameters $\epsilon_v$.

Otherwise, if the input contains more than one image, \emph{i.e.} $N>1$, DFM will first concatenate the features of those inputs by
\begin{equation}
\mathbf{\widetilde{F}}^i =  \operatorname{Cat}(\mathbf{F}_1^i, \mathbf{F}_2^i, .., \mathbf{F}_N^i),
\end{equation}
where $\mathbf{\widetilde{F}}^i$ denotes the concatenated features. $\operatorname{Cat}(*)$ denotes the concatenation operation.
It should be noted that, unlike previous concatenation schemes, which often concatenate the input feature along the channel dimension, our scheme concatenates in a new dimension, namely the modality dimension. Accordingly, the size of the concatenated features
$\mathbf{\widetilde{F}}^i$ is $N \times C\times H \times W$. \textcolor{blue}{It should be noted that such an operation can ensure that the orders of those modalities within the inputs can be arbitrary without affecting their outputs. }

Then, three $1\times 1$ convolutional layers are employed to project $\mathbf{\widetilde{F}}^i$ into a key feature space, a query feature space, and a value feature space, respectively \emph{i.e.}
\begin{equation}
\begin{split}
	&\mathbf{K}^i =  \operatorname{Conv}(\mathbf{\widetilde{F}}^i; \epsilon_k ), \\&
	\mathbf{Q}^i =  \operatorname{Conv}(\mathbf{\widetilde{F}}^i; \epsilon_q), \\&
	\mathbf{V}^i =  \operatorname{Conv}(\mathbf{\widetilde{F}}^i; \epsilon_v),
\end{split}
\end{equation}
where $\operatorname{Conv}(*; \epsilon_k)$, $\operatorname{Conv}(*; \epsilon_q)$ and  $\operatorname{Conv}(*; \epsilon_v)$ denote three $1\times 1$ convolutional layers with their parameters $\epsilon_k$, $\epsilon_q$  and $\epsilon_v$, respectively. $\mathbf{K}^i$, $\mathbf{Q}^i$ and $\mathbf{V}^i \in \mathbb{R}^{N \times C\times H \times W}$
denote the key features, query features and value features, respectively. Note that, the operation $\operatorname{Conv}(*; \epsilon_v)$ is also applied to the input features when there is only one image as the input.

Then, DFM explores the cross-modal relations across each feature channel, respectively, among different modalities by 
\begin{equation}
\begin{split}
	&\mathbf{\hat{K}}^{i}_{c} = \operatorname{Reshape}(\mathbf{K}^i_{c}), \\ &
	\mathbf{\hat{Q}}^{i}_{c} = \operatorname{Reshape}(\mathbf{Q}^i_{c}), \\ &
	\mathbf{W}^{i}_{c}=   \operatorname{Softmax} \left (\frac{\mathbf{\hat{K}}^{i}_{c} \odot (\mathbf{\hat{Q}}^{i}_{c})^T }{\sqrt{N}} \right ),
\end{split}
\end{equation}
where $c=1,2,...,C$ denotes the $c$-th channel of the key/query features. $\odot$ denotes the matrix multiplication. $\operatorname{Reshape}(*)$ denotes the operation that changes the input features' size from $\mathbb{R}^{N \times 1 \times H \times W}$ to $\mathbb{R}^{N\times HW}$. $(*)^T$ denotes the transpose operation. $ \operatorname{Softmax}(*) $ denotes the Softmax function. Accordingly, $\mathbf{W}^{i}_{c} \in \mathbb{R}^{N\times N} $ builds the relations among the $c$-th channel of $N$ types of unimodal features. For example, the first element in the first row of  $\mathbf{W}^{i}_{c} $ indicates the relations between the $c$-th feature channel of the first modality and itself. While the second element in the first row of $\mathbf{W}^{i}_{c} $ denotes the relations between the $c$-th feature channel of the first modality and the $c$-th feature channel of the second modality, and so on. 

Next, DFM explores such interactions among the unimodal features of $N$ input images to enhance themselves  by using the information of other modalities,  \emph{i.e.}
\begin{equation}
\begin{split}
	\mathbf{F}^{i}_{E,c}= \operatorname{R\_Reshape(\mathbf{W}^{i}_{c} \odot \operatorname{Reshape}(\mathbf{V}^{i}_{c}))} ,
\end{split}
\end{equation}
where $\mathbf{F}^{i}_{E,c} \in \mathbb{R}^{N \times 1 \times H \times W}$ denotes the $c$-the channel of the enhanced  features.  $\operatorname{R\_Reshape(*)}$ denotes the reverse operation of $\operatorname{Reshape}(*)$, which will change the input features' size  from $\mathbb{R}^{N\times HW}$ to $\mathbb{R}^{N \times 1 \times H \times W}$. Furthermore, all the $C$ channels of unimodal features of $N$ input images are obtained in the same way, thus obtaining their corresponding enhanced features $\mathbf{F}^{i}_{E} \in \mathbb{R}^{N \times C \times H \times W}$. 

Finally, those enhanced unimodal features are fused to capture their cross-modal complementary information by 
\begin{equation}
\mathbf{F}^{i}_{f} = \operatorname{Mean}_N(\mathbf{F}^{i}_{E}),
\end{equation}
where the features $\mathbf{F}^{i}_{f} \in \mathbb{R}^{ C \times H \times W} $ denote the $i$-th level of fused features. $\operatorname{Mean}_N(*)$ denotes the average operation along the modality dimension for the features  of $N$ modalities. \textcolor{blue}{This operation will ensure the value ranges and sizes of corresponding outputs are the same for different inputs.}

As shown in Fig. \ref{fig_Frame}, for each level of unimodal features, an DFM is employed to fuse those unimodal features for extracting their cross-modal complementary information. Totally, we may obtain  five levels of fused features, \emph{i.e.} $\mathbf{F}^{1}_{f}, \mathbf{F}^{2}_{f}, \mathbf{F}^{3}_{f}, \mathbf{F}^{4}_{f}$ and $\mathbf{F}^{5}_{f}$.

\textcolor{blue}{Overall, the proposed DFM eliminates the necessity of modifying the model structure and barely introducing any additional parameters when facing the changes in the number of modalities within the inputs. Moreover, its outputs are irrelevant to the orders of those modalities within the inputs. And, the value range and sizes of its outputs are the same for all the types of inputs. }
Therefore, by virtue of our proposed DFM, our proposed model can adaptively establish relations among different unimodal features and effectively capture multi-modal complementary information from any number of inputs with arbitrary modalities, achieving the AM SOD.



\subsection{Saliency prediction decoder (SPD)}

After obtaining different levels of fused features $\mathbf{F}^{i}_{f}$, the next step is to exploit them for detecting those salient objects. To this end, a saliency prediction decoder (SPD) is designed in this paper. 

\begin{figure}[!t]
	\centering
	\includegraphics[width=0.8\linewidth]{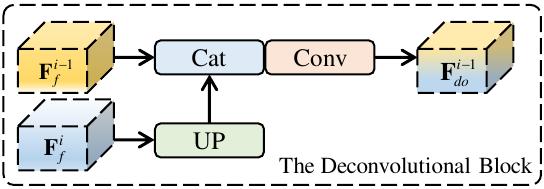}
	\caption{Basic structure of the deconvolutional block.}
	\label{fig_Decoder}
\end{figure}

Specifically, as shown in Fig. \ref{fig_Frame}, SPD follows the classical coarse-to-fine structure and employs five deconvolutional blocks to predict the saliency maps gradually. The basic structure of those deconvolutional blocks is shown in Fig. \ref{fig_Decoder}.  It first upsamples output features $\mathbf{F}^{i}_{do}$ from the $i$-th level of the deconvolutional block to the size of the fused features $\mathbf{F}^{i-1}_{f}$ from the previous level (if existed) by using a bilinear interpolation algorithm. Then, the upsampled features will be concatenated with their previous level of fused features. After that, a convolutional layer is employed to exploit their cross-level complementary information, thus obtaining the $(i-1)$-th level of output features. Mathematically, this process can be expressed by
\begin{equation}
\mathbf{F}^{i-1}_{do} =  \operatorname{Conv}( \operatorname{Cat}(\mathbf{F}^{i-1}_{f}, \operatorname{UP}(\mathbf{F}^{i}_{do})); \nu_{i-1}),
\end{equation}
where $\operatorname{Conv}(*; \nu_{i-1})$ denotes a convolutional layer with its parameters $ \nu_{i-1}$. $\operatorname{UP}(*)$ denotes the upsample operation.
Accordingly, four levels of output features $\mathbf{F}^{1}_{do}$, $\mathbf{F}^{2}_{do}$, $\mathbf{F}^{3}_{do}$ and $\mathbf{F}^{4}_{do}$ are obtained.

Finally, a convolutional layer is employed for each level of output features to predict the corresponding saliency map, \emph{i.e.}
\begin{equation}
\mathbf{S}^{i} =  \operatorname{Conv}( \mathbf{F}^{i}_{do}; \nu_p^i),
\end{equation}
where $\mathbf{S}^{i}$ is the $i$-th level of predicted saliency map. $\operatorname{Conv}(*; \nu_p^i)$ denotes a convolutional layer with its parameters $\nu_p^i$ for saliency prediction. It should be noted that, in the testing stage, $\mathbf{S}^1$ is used as the final saliency map. 

\subsection{Loss function}
We use the cross-entropy loss to train our proposed model, which is performed on each predicted saliency map, \emph{i.e.}
\begin{equation}
L_{ce}= \sum_{i=1}^{4} \left( \mathbf{Y}^i\log (\mathbf{S}^i)+(1-\mathbf{Y}^i)\log (1-\mathbf{S}^i) \right),
\end{equation}
where $\mathbf{Y}^i$ denotes the ground truth for the $i$-th level of predicted map. Accordingly, the total loss for our model is formulated as:
\begin{equation}
	L_{total}= L_{modality} + L_{ce}.
\end{equation} Here, $L_{modality}$, which is defined by Eq. (7), is responsible for supervising the classifier training.

\section{Experiments} \label{sec::ex}

\subsection{Datasets and Evaluation Metrics}

A new dataset, AM-XD dataset, is constructed for our proposed AM SOD on top of some existing SOD datasets (shown in Table \ref{tab_110}), including: RGB SOD dataset(DUTS \cite{DUTS}), RGB-D SOD datasets (NJU2K\cite{NJU}, NLPR\cite{NLPR}, DUT-RGBD\cite{r33}, SIP \cite{SIP} and STEREO\cite{STEREO}), RGB-T SOD(VT821\cite{VT821}, VT1K\cite{VT1K} and VT5K\cite{VT5K}), and RGB-D-T dataset(VDT-2048\cite{9931143}).

\subsubsection{Training set} As shown in Table \ref{tab_110}, to build our AM-XD's training set, we first randomly sample 5000 images from DUTS-TR rather than all 10553 images, considering that there are too many RGB samples. Then, we obtain  1485, 700 and 800 RGB-D images, respectively, from the three RGB-D SOD datasets. Moreover, for VT5K\cite{VT5K}  and VDT-2048 \cite{9931143}, we take their default setting and obtain 2500 RGB-T images and 1048 RGB-D-T images, respectively. Accordingly, our training set have total 11533 samples, including 5000 RGB SOD images, 2985 RGB-D SOD image pairs, 2500  RGB-T SOD image pairs and 1048 RGB-D-T image sets.

\begin{figure}[!t]
	\centering
	\includegraphics[width=0.95\linewidth]{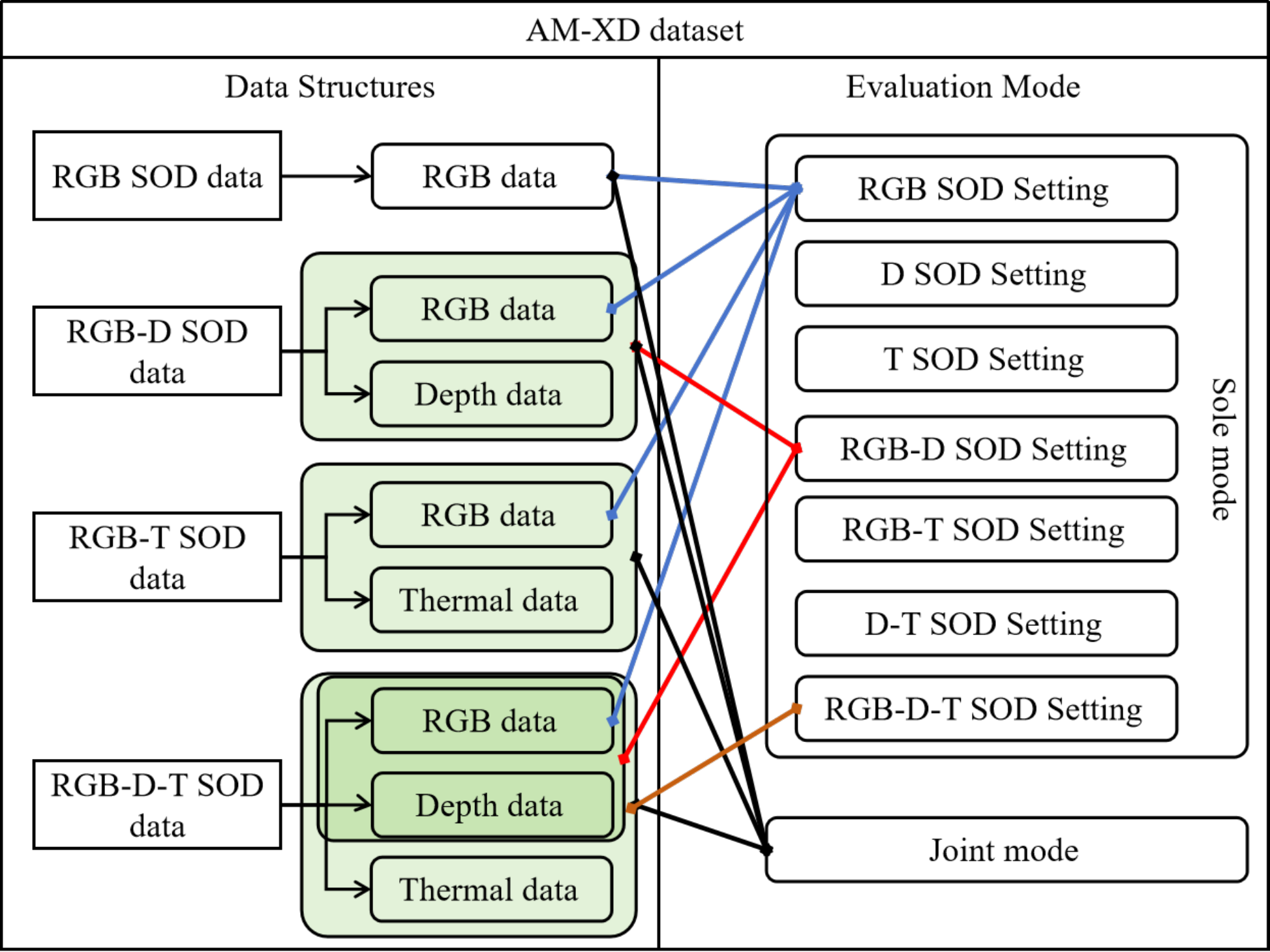}
	\caption{Data structures and evaluation mode of our proposed AM-XD dataset. Here, we only draw the connection lines of RGB SOD setting, RGB-D SOD setting, RGB-D-T SOD setting of sole mode and joint mode for better understanding.}
	\label{fig_datac}
\end{figure}

\subsubsection{Testing set}  As shown in Table \ref{tab_110},  our AM-XD's testing set contains  all the samples from the testing sets of  DUTS\cite{DUTS}, NJU2K \cite{NJU}, NLPR \cite{NLPR}, DUT-RGBD\cite{r33}, VT5K\cite{VT5K} and VDT-2048 \cite{9931143} datasets, and all the samples of SIP \cite{SIP}, STEREO \cite{STEREO},  VT821\cite{VT821} and VT1K\cite{VT1K} datasets. Accordingly, our testing set contains total 13442 samples, including 5000 RGB images, 3121 RGB-D image pairs and 4321 RGB-T image pairs and 1000 RGB-D-T image sets.

\begin{table*}[!t]
	\renewcommand{\arraystretch}{1.3}
	\caption{Training and Testing sets of our proposed AM-XD dataset.} 
	\label{tab_110}
	\centering
		\resizebox{\textwidth}{!}{
		\begin{tabular}{c|c|ccccc|ccc|c}
			\hline
			&\multicolumn{10}{c }{Training Set}  \\
			\hline
			Modalities & RGB SOD& \multicolumn{5}{c|}{RGB-D SOD} &  \multicolumn{3}{c|}{RGB-T SOD}  & RGB-D-T SOD  \\
			\hline
			Dataset & DUTS-TR \cite{DUTS}  & NJU2K \cite{NJU} & NLPR \cite{NLPR} & DUT-RGBD \cite{r33} & & & \multicolumn{3}{c|}{VT5K\cite{VT5K}00}   & VDT-2048 \cite{9931143}   \\ 
			\hline
			Number  & 5000 & 1485  &700   &800  & & &\multicolumn{3}{c|}{2500}  &  1048   \\ 
			\hline
			Total & 5000& \multicolumn{5}{c|}{2985}   &  \multicolumn{3}{c|}{2500}   & 1048   \\
			\hline
			\hline
			&\multicolumn{10}{c}{Testing Set} \\
			\hline
		Modalities	& RGB SOD& \multicolumn{5}{c|}{RGB-D SOD}& \multicolumn{3}{c|}{RGB-T SOD}&RGB-D-T SOD \\
			\hline
Dataset &  DUTS-TE \cite{DUTS}  & NJU2K \cite{NJU} & NLPR \cite{NLPR}  & SIP \cite{SIP} & DUT-RGBD \cite{r33}  & STEREO \cite{STEREO} & VT821\cite{VT821} & VT1K\cite{VT1K} & VT5K\cite{VT5K} & VDT-2048 \cite{9931143}\\
\hline
Number &  5000 &  500&  300&  921&  400&  1000& 821&1000& 2500&   1000\\
\hline
Total & 5000& \multicolumn{5}{c|}{3121}& \multicolumn{3}{c|}{4321 }& 1000\\
\hline
		\end{tabular}}
\end{table*}

\begin{table*}[!t]
\renewcommand{\arraystretch}{1.3}
\caption{Testing modes of our proposed AM-XD dataset.} 
\label{tab_888}
\centering
\begin{tabular}{c|ccc|ccc|c|c}
	\hline 
	\multirow{2}{*}{Mode} & \multicolumn{7}{c|}{Sole mode}& Joint mode \\
	\cline{2-9}
	& \multicolumn{3}{c|}{Single-modality SOD} & \multicolumn{3}{c|}{Two-modality SOD}& \multicolumn{1}{c|}{Three-modality SOD}& ALL\\
	\hline 
	Modalities & RGB & D & T & RGB-D & RGB-T & D-T & RGB-D-T &   - \\
	\hline 
	Number & 13466 & 4145 & 5345 & 4121 & 5321 & 1000 & 1000 &   -\\
	\hline 
\end{tabular}
\end{table*}

\subsubsection{Evaluation modes} 

\textcolor{blue}{As shown in Table \ref{tab_888} and Fig. \ref{fig_datac}, we set two evaluation modes for our proposed AM-XD SOD dataset, \emph{i.e.} sole mode and joint mode. Under sole mode, there are seven settings: RGB SOD setting, depth (D) SOD setting, thermal (T) SOD setting, RGB-D SOD setting, RGB-T SOD setting, D-T SOD setting, and RGB-D-T SOD setting. In the RGB setting, we will take all the RGB images within the AM-XD's testing set as the inputs and evaluate AM SOD models' outputs using ground truths. Other test settings under the sole mode follow a similar procedure. Furthermore, the joint mode of our proposed AM-XD dataset takes  the RGB data from RGB SOD dataset, RGB-D data from RGB-D SOD dataset, RGB-T data from RGB-T SOD dataset, and RGB-D-T data from RGB-D-T SOD dataset as the inputs of different AM SOD model for evaluation. 
}

\subsection{Evaluation Metrics}

We utilize four common metrics to assess our proposed model's performance: mean absolute error ($\mathcal{M}$)\cite{NJU}, mean F-measure ($F_\beta$)\cite{NJU}, mean S-measure ($S_\alpha$) \cite{r25} and mean E-measure ($E_\gamma$) \cite{r26}. 

Mean absolute error ($\mathcal{M}$) measures the average absolute difference between the predicted saliency map $\mathbf{S}$ and the ground truth $\mathbf{Y}$, \emph{i.e.}
\begin{equation}\label{eq_15}
\mathcal{M} = \frac{1}{W \times H} \sum_{x=1}^{W} \sum_{y=1}^{H}|\mathbf{S}(x,y)-\mathbf{Y}(x,y)|,
\end{equation}
where $W$ and $H$ represent the width and height of the saliency map (or ground truth), respectively.

F-measure ($F_\beta$) is also a commonly used metric in SOD, which provides a balanced measure for SOD models' accuracy by  combining precision and recall into a single score, which is expressed by
\begin{equation}\label{eq14}
F_{\beta} = \frac{(1+\omega^2) \times Precision \times Recall}{\omega^2 \times Precision + Recall}.
\end{equation}
Here, $Precision$ computes the proportion of true positive predictions out of all positive predictions and  $Recall$  represents the proportion of true positive predictions out of all actual positive instances.  
Following \cite{NJU}, we set $\omega^2$ to 0.3.

\subsection{Implementation details}

\subsubsection{Training procedure}

\textcolor{blue}{We use the widely used Pytorch library to compile our proposed model and run it on an NVIDIA 2080Ti GPU. We initialize the parameters of our proposed MSFE on top of a ResNet50\cite{ResNet50} network pre-trained on Imagenet\cite{ImageNet}. While we initialize the parameters of other modules by using the Kaiming initialization\cite{KaiMingInit}. We employ the SGD algorithm\cite{SGD} with Nesterov momentum to train our proposed model by setting its learning rate to 2e-3 and weight decay to 5e-4. 
Furthermore, we dynamically reduce its learning rate by multiplying 0.8 for every 20 epochs. }

\textcolor{blue}{The training set of our proposed AM-XD dataset simultaneously contains RGB data, depth data, thermal data, RGB-D data, RGB-T data, and RGB-D-T data. In the training, we will first randomly sample a batch of RGB data, depth data, thermal data, RGB-D data, RGB-T data, or RGB-D-T data. Then, we will resize the sizes of those samples to  $224 \times 224$ and augment them by using random cropping and flipping. After that, we will feed them into our proposed MSN and obtain their corresponding saliency maps. Finally, we will compute the loss between those predicted saliency maps and ground truths for optimization. }

\subsubsection{Testing procedure}

\textcolor{blue}{The proposed MSN is trained once on the training set of our AM-XD dataset. After training, we evaluate it under different test settings of AM-XD dataset. }

\begin{table*}[!t]
	\renewcommand{\arraystretch}{1.3}
	\caption{Quantitative results of different models on AM-XD SOD dataset.} 
	\label{tab_88}
	\centering
	\resizebox{\textwidth}{!}{
		\begin{tabular}{cc|cccccc|cccccc|cc|cc}
			\hline 
			& & \multicolumn{14}{c|}{Sole Mode} & \multicolumn{2}{c}{Joint Mode} \\
			\hline
			& & \multicolumn{6}{c|}{Single-modality SOD} & \multicolumn{6}{c|}{Two-modality SOD}& \multicolumn{2}{c|}{Three-modality SOD}& \multicolumn{2}{c}{ } \\
			\hline 
			& \multirow{2}{*}{ Models } & \multicolumn{2}{c|}{RGB} & \multicolumn{2}{c|}{D}  & \multicolumn{2}{c|}{T} & \multicolumn{2}{c|}{RGB-D} & \multicolumn{2}{c|}{RGB-T}& \multicolumn{2}{c|}{D-T}& \multicolumn{2}{c|}{RGB-D-T} & \multicolumn{2}{c}{ALL} \\
			\cline{3-18}
			&  &$\mathcal{M}\downarrow$ &$F_\beta\uparrow$ &$\mathcal{M}\downarrow$ &$F_\beta\uparrow$ &$\mathcal{M}\downarrow$ &$F_\beta\uparrow$ &$\mathcal{M}\downarrow$ &$F_\beta\uparrow$ &$\mathcal{M}\downarrow$ &$F_\beta\uparrow$ &$\mathcal{M}\downarrow$ &$F_\beta\uparrow$ &$\mathcal{M}\downarrow$ &$F_\beta\uparrow$&$\mathcal{M}\downarrow$ &$F_\beta\uparrow$ \\
			\hline 
			&  PFSNet(2019) \cite{o12} &0.045&0.810&0.099&0.559&0.063&0.629 &- & - &- &- &-& -&- &- &- &-    \\ 
			RGB&  PSGLoss(2021) \cite{o14}&0.047&0.722&0.099&0.482&0.070&0.507 &- & - &- &- &-& -&- &- &- &-  \\ 
			SOD&  PoolNet++(2023) \cite{RGB2} &0.041&0.822&0.120&0.450&0.810&0.458&- & - &- &- &-& -&- &- &- &-  \\ 
			&\textcolor{blue}{LeNo(2023) \cite{LeNo} } &\textcolor{blue}{0.043}&\textcolor{blue}{0.814}&\textcolor{blue}{0.110}&\textcolor{blue}{0.514}&\textcolor{blue}{0.078}&\textcolor{blue}{0.516}&- &- &- &- &- &- &- &- &- &- \\  
			&  SefReFormer(2023) \cite{o15}& \textbf{0.033} &\textbf{0.844}    & 0.098    
			&0.576 &0.066    & 0.717 &- & - &- &- &-& -&- &-&- &-  \\  
			\hline  
			& MobileSal$^\dagger$(2021)\cite{n73}&- &- &- &- &- &- &0.048&0.751&0.067&0.624&0.020&0.480&- &- &- &-   \\ 
			& MLF$^\dagger$(2022)\cite{k02} &- &- &- &- &- &- &0.480&0.762&0.110&0.524&0.045&0.410&- &- &- &-  
			\\    
			RGB-D & VST$^\ddagger$(2022)\cite{n33} &0.044&0.817&0.093&0.629&0.083&0.682 &0.038&0.816&0.035&0.803&0.074&0.493&- &- &- &-  \\ 
		SOD	&  SwinNet$^\ddagger$(2022) \cite{n30} &- &- &- &- &- &- &0.045&0.786&\textbf{0.028}&\textbf{0.837}&0.027&0.518&- &- &- &- \\
			& CAVER$^\ddagger$(2023)\cite{z27}  &- &- &- &- &- &- &0.044&0.715&0.038&0.760&0.089&0.469&- &- &- &-\\
			&  \textcolor{blue}{CPNet$^\dagger$(2024)} \cite{CMFPD}&- &- &- &- &- &- &\textcolor{blue}{\textbf{0.033}}&\textcolor{blue}{0.822}&\textcolor{blue}{0.028}&\textcolor{blue}{0.811}&\textcolor{blue}{0.020}&\textcolor{blue}{0.320}&- &- &- &- \\ 
			\hline 
			& DCNet$^\dagger$(2022) \cite{RGBTSOD4} &- &- &- &- &- &- &0.098&0.635&0.059&0.702&0.120&0.327&- &- &- &-   \\
			&  APNet$^\dagger$(2022) \cite{z14}&- &- &- &- &- &- &0.066&0.695&0.036&0.743&0.036&0.422&- &- &- &-    \\
			RGB-T& TNNet$^\dagger$(2023)\cite{z28} &- &- &- &- &- &- &0.074&0.698&0.049&0.741&0.110&0.298&- &- &- &-     \\	
			SOD& FANet$^\dagger$(2023)\cite{z29} &- &- &- &- &- &- &0.067&0.712&0.035&0.767&0.061&0.382&- &- &- &-  
			\\
			& LSNet$^\dagger$(2023)\cite{z25} &- &- &- &- &- &- &0.064&0.714&0.060&0.676&0.064&0.390&- &- &- &-  \\	
			\hline  
			RGB-D-T & HWSI (2023)\cite{9931143} &- &- &- &- &- &- &- &- &- &- &- &-&\textbf{0.0026} & \textbf{0.896} &- &-    \\
			SOD& MFFNet (2023)\cite{10171982} &- &- &- &- &- &- &- &- &- &- &- &-&0.0032 &  0.871 &- &-   \\ 
			\hline 
			&  OUR & 0.055  &0.803 &\textbf{0.064} &\textbf{0.700} & \textbf{0.044} & \textbf{0.769}   &   {0.035} &\textbf{0.850} &0.038 &0.833 &\textbf{0.0065} & \textbf{0.740} &0.0035 &  0.841  &0.049     &0.816\\ 
			\hline
		\end{tabular}
	}
\end{table*}

\begin{table*}[!t]
	\renewcommand{\arraystretch}{1.3}
	\caption{Quantitative results of each component of our proposed model.} 
	\label{tab_98}
	\centering
	\resizebox{\textwidth}{!}{
		\begin{tabular}{c|c|cc|cc|cc|cc|cc|cc|cc|cc}
			\hline 
			\multirow{3}{*}{Models}	& \multirow{2}{*}{Params}& \multicolumn{14}{c|}{Sole} &   \multicolumn{2}{c}{Joint} \\		\cline{3-18} 
			& & \multicolumn{2}{c|}{RGB} & \multicolumn{2}{c|}{D}  & \multicolumn{2}{c|}{T} & \multicolumn{2}{c|}{RGB-D} & \multicolumn{2}{c|}{RGB-T}& \multicolumn{2}{c|}{D-T}& \multicolumn{2}{c|}{RGB-D-T} & \multicolumn{2}{ c}{ALL} \\
			\cline{2-18} 
			&M &$\mathcal{M}\downarrow$ &$F_\beta\uparrow$ &$\mathcal{M}\downarrow$ &$F_\beta\uparrow$ &$\mathcal{M}\downarrow$ &$F_\beta\uparrow$ &$\mathcal{M}\downarrow$ &$F_\beta\uparrow$ &$\mathcal{M}\downarrow$ &$F_\beta\uparrow$ &$\mathcal{M}\downarrow$ &$F_\beta\uparrow$ &$\mathcal{M}\downarrow$ &$F_\beta\uparrow$&$\mathcal{M}\downarrow$ &$F_\beta\uparrow$ \\
			\hline 
			Baseline & \textbf{43.87} &0.089&0.71&0.096&0.635&0.059&0.724&0.075&0.780&0.056&0.797&0.0083&0.702&0.0054&0.801&0.081&0.763   \\
			Baseline+MS  & 90.89
			&\textbf{0.048}&\textbf{0.811}&\textbf{0.069}&0.694&0.047&0.768&0.045&0.812&\textbf{0.037}&0.803&0.0110&0.661&0.0057&0.766&0.056&0.783
			 \\
			+MSFE	& 46.65 	&0.053&0.787&0.070&0.687&0.045&0.761&0.059&0.804&0.044&0.818&0.0060&0.714&0.0042&0.808&0.068&0.772\\
			+MSFE+DFM  	& 47.29&0.055  &0.803 &0.064 &\textbf{0.700} & \textbf{0.044} & \textbf{0.769}   &  \textbf{0.035} &\textbf{0.850} &0.038 &\textbf{0.833} &\textbf{0.0065} & \textbf{0.740} &\textbf{0.0035} &  \textbf{0.841}  &\textbf{0.049}     &\textbf{0.816}  \\  
			\hline
	\end{tabular}}
\end{table*}

\subsection{Comparison with SOTA models}

We compare our proposed AM model with several state-of-the-art SOD models, including \textbf{RGB SOD models}( PFSNet \cite{o12},  PoolNet++ \cite{RGB2}, PSGLoss \cite{o14} and SefReFormer \cite{o15}), \textbf{RGB-D SOD models} (MobileSal$^\dagger$(2021),   MLF$^\dagger$(2022)\cite{k02},  VST$^\ddagger$(2022)\cite{n33},  SwinNet$^\ddagger$(2022) \cite{n30},  CAVER$^\ddagger$(2023)\cite{z27}), \textbf{RGB-T SOD models} (DCNet$^\dagger$(2022) \cite{RGBTSOD4},   APNet$^\dagger$(2022) \cite{z14}, TNNet$^\dagger$(2023)\cite{z28}, FANet$^\dagger$(2023)\cite{z29} and LSNet$^\dagger$(2023)\cite{z25}), and \textbf{RGB-D-T SOD models}(HWSI(2023)\cite{9931143} and MFFNet(2023)\cite{10171982}). 
It should be noted that some authors separately train their  models on the RGB-T SOD datasets and RGB-D SOD datasets. We mark these models by  the $^\ddagger$. Accordingly, the models marked by $^\dagger$ are only designed for RGB-D or RGB-T SOD tasks.  Besides, VST \cite{n33} has an RGB SOD version, and we also give its RGB SOD results.  For fair comparisons, we obtain their codes and their trained parameters from the authors and directly test those models on our dataset. All results are shown in Table \ref{tab_88}. It should be noted that the $MAE$ vaules of D-T and RGB-D-T settings are notably lower than other settings due to the fact that salient objects within the data of such two settings are small one.

From Table \ref{tab_88}, we may easily draw the following conclusions. 
First, directly applying existing models to inputs that fall outside their target modalities may result in a significant performance drop.  
For example, SelfReFormer$^\dagger$(2023)\cite{o15} achieves much better results in the RGB setting compared to those in the D or T settings, since it is specially trained for RGB SOD.
Similarly, FANet$^\dagger$(2023)\cite{z29} obtains superior results for RGB-T SOD setting than those for RGB-D setting or D-T setting, since it is trained for RGB-T SOD.
Secondly, limited by their structures, existing SOD models cannot well handle the inputs that contain more or less images than their design settings.
For instance, those RGB SOD models cannot detect salient objects from RGB-T, RGB-D, or RGB-T-D images, and those RGB-D/RGB-T models are also inappropriate for RGB SOD or RGB-D-T SOD inputs.
Different from existing SOD methods, our proposed MSN can detect salient objects with considerable performance across inputs of arbitrary modality types and numbers
after only one training. 
This is primarily owes to the proposed MSFE, which enables our network to effectively  extract arbitrary modality information well with the assistance of  modality indicators.
Additionally, it is also attributed to the proposed DFM, which allows our network to dynamically fuse any amount of modality information for salient object detection.

However, we may also find that, for single-modality SOD, our proposed model achieves 0.055 in $\mathcal{M}$ and 0.803 in  $F_\beta$ under the RGB SOD setting. Such results are inferior to most existing SOTA RGB SOD models. But, our proposed MSN achieves the best results in the settings of D SOD and T SOD, respectively. 
These results are kindly reasonable. First, existing RGB SOD models are not designed for depth images or thermal images, thus failing to detect salient objects from those modalities. Secondly, compared with existing SOD models, which just detect salient objects from one particular type of inputs, our proposed AM model has learned to simultaneously detect those salient objects from Arbitrary Modalities. 
Despite that, the significant discrepancies among different modality data hinder the network from simultaneously achieving the best fit to  their data distributions. Therefore, the results obtained by our method are inferior to those achieved by those models specifically designed for RGB modality.
Such issue will be prioritized in our future work.	

For multi-modality SOD, our proposed MSN achieves the best results under RGB-D SOD and D-T SOD settings and competitive results for RGB-T SOD and RGB-D-T settings. Moreover, it achieves  0.049 in $\mathcal{M}$ and 0.816 in  $F_\beta$ under ALL setting. This proves the feasibility of our proposed AM SOD task and the effectiveness of our proposed model. Predictably, AM models will eventually achieve more competitive results than those SOTA SOD models with the development of relevant theories and technological progress.


\subsection{Ablation study}
 
\subsubsection{Effectiveness of each component of our proposed model}

As shown in Table \ref{tab_98}, several variants of our proposed model are designed to validate the effectiveness of each component in our proposed model. Specifically, `Baseline' is the model that removes the proposed MSFE and DFM from our proposed model. It uses the siamese ResNet50\cite{ResNet50} as the backbone and the element-wise addition for feature fusion.
`Baseline+MS' is the model that employs multi-stream feature extractors for different modality feature extraction.
`Baseline+MSFE' is the model that uses our proposed MSFE as the backbone and `Baseline+MSFE+DFM' is the model that further employs our proposed DFM for feature fusion.

As shown in Table \ref{tab_98}, compared with the performance of `Baseline', desirable performance improvements can be achieved by using `Baseline+MS'. 
Such observations is consistent with our previous conclusion that the discrepancies among different modalities will prevent an AM SOD model from accurately fitting the distributions of all modalities simultaneously via one feature extractor, leading to suboptimal results.
Although the desirable improvements are achieved by `Baseline+MS', it is obvious that the total parameters will be also significantly increased if we design one feature extractor for one modality.  This is undesirable for those resource-limited devices.
Differently, our proposed MSFE can also significantly boost the performance of `Baseline' on the three datasets while requiring fewer additional parameter costs. 
This may owe to the input modality indicators, which enables our network to adaptively extract those discriminative unimodal features from the inputs by using one feature extractor, thus improving performance. 
Furthermore, as shown in Table \ref{tab_98}, the performance gains of our proposed DFM on the setting of single-modality SOD are little, since single-modality SOD does not involve multi-modal feature fusion.
While, on the other two datasets, our proposed DFM gains large improvements on all metrics.
The reason behind this is that our proposed DFM can effectively exploit the cross-modality complementary information from multi-modal inputs, thus boosting performance.


 \begin{table*}[!t]
	\renewcommand{\arraystretch}{1.3}
	\caption{Quantitative results of our proposed MSN with different types of inputs} 
	\label{tab_100}
	\centering
	\begin{tabular}{cccccccc}
		\hline 
		& RGB& D & T & RGB-D &RGB-T& D-T& RGB-D-T  \\
		\hline 
		$\mathcal{M}\downarrow$ &0.0062   &0.0190  &0.0078  & 0.0046  &0.0036  &0.0056  &\textbf{0.0035 }   \\  
		$F_\beta\uparrow$ 	& 0.7653 & 0.4396  & 0.7386 & 0.7947  &0.8383 &0.7463  &\textbf{0.8410}    \\  
		\hline
	\end{tabular}
\end{table*}

\subsubsection{Quantitative comparisons with our proposed MSN for different types of inputs}

\textcolor{blue}{As shown in Table \ref{tab_100}, we will further display the results of using different number of modalities for the same scene. For that, we first collect all the RGB-D-T data within our proposed AM-SOD dataset. Then, we only employ the parts of RGB data, depth data, thermal data, RGB-D data, RGB-T data, D-T data and RGB-D-T data, respectively, within those RGB-D-T data to test our proposed MSN. By doing so, we can estimate our proposed model under different types of inputs with the same scene.  }

It can be seen that, when using the images of one modality for detection, `Ours (RGB)' obtains better results than `Ours (T)' and `Ours (D)'. This may result from the fact that RGB images usually contain more information than the other two modalities, thus having more diverse saliency patterns.
Furthermore, it can be also observed that our model's performance will be constantly better by adding more information of different modalities, \emph{i.e.} the performance of our dual-modality based models  (`Ours (RGB-D)', `Ours (RGB-T)' and `Ours (D-T)') are better than those single-modality based ones (`Ours (RGB)', `Ours (T)' and `Ours (D)'), but inferior than the triple-modality based one (`Ours (RGB-D-T)'). This proves that exploiting the complementary information within multi-modal images can make up for the drawbacks of single-modality images and provide more scene information, thus helping to accurately detect those salient objects. 
 
However, as shown in Table \ref{tab_100}, our proposed MSN for the inputs with less modalities may obtain very close results to those with more modalities in some specific settings, \emph{e.g.} RGB vs D-T or RGB-T vs RGB-D-T.
As we know, using more modalities will also significantly increase the computational complexity and inference times. Taking our proposed model as an example, as shown in Table \ref{tab_11}, it can be seen that using one more modality will increase about 35\% FLOPS and reduce about 31\% FPS. 
This also verifies our viewpoint and the importance of our proposed AM SOD task, \emph{i.e.} it is not necessary to use all the equipped cameras of different modalities for detection in real-life applications even though our devices may have multiple cameras.

 \begin{table}[!t]
 	\renewcommand{\arraystretch}{1.3}
 	\caption{Efficiency and parameters.} 
 	\label{tab_11}
 	\centering
 	\begin{tabular}{cc|c|c}
 		\hline
 		Models & RGB/D/T & RGB-D/RGB-T/D-T & RGB-D-T \\ \hline
 		FLOPS(G) & 35.49 & 55.4 &  73.55 \\ \hline
 		Speed(FPS) & 40.8 & 25.6 & 17.6 \\ \hline
 		Params(M) &\multicolumn{3}{c}{ 47.29 }  \\ 
 		\hline
 	\end{tabular}
 \end{table}

\begin{figure*}[!t]
	\centering
	\includegraphics[width=0.75\linewidth]{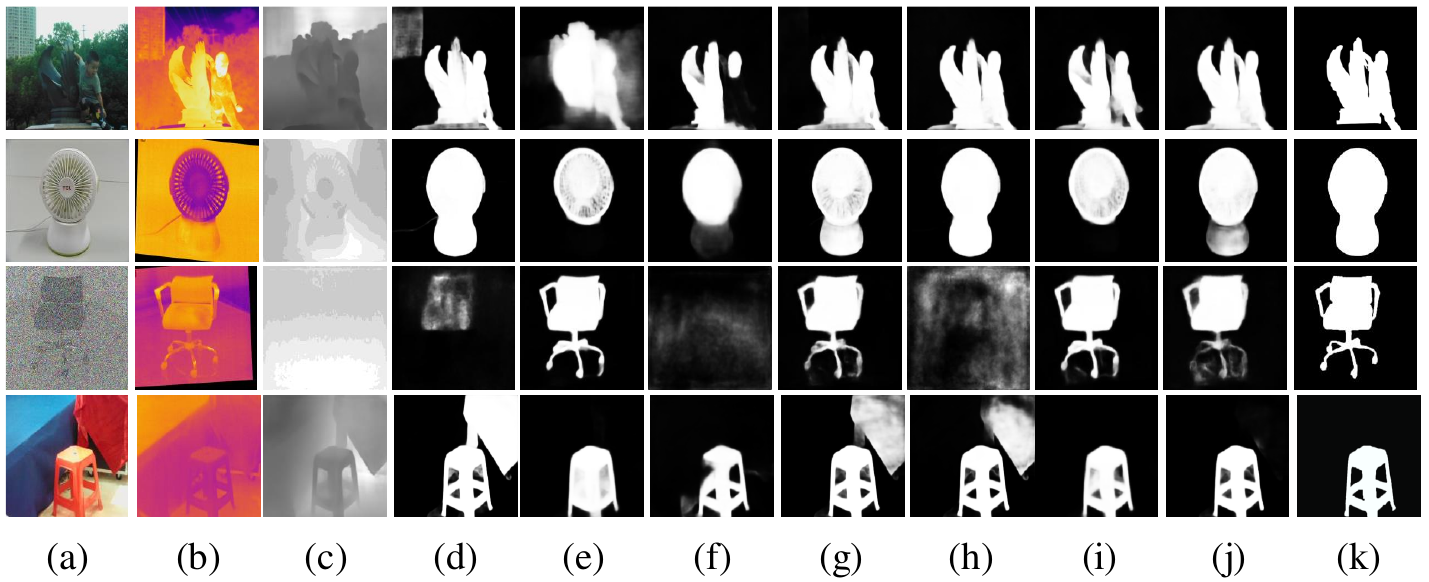}
	\caption{Saliency maps detected by uisng different inputs. (a) RGB images; (b) Thermal images; (c) Depth images; 
	(d) Saliency maps detected from RGB images;
	(e) Saliency maps detected from thermal images;
	(f) Saliency maps detected from depth images;
	(g) Saliency maps detected from RGB-T images;
	(h) Saliency maps detected from RGB-D images;
	(i) Saliency maps detected from D-T images;
	(j) Saliency maps detected from RGB-D-T images; (k) Ground Truth.   }
	\label{fig_100}
\end{figure*}

  
\begin{figure}[!t]    
	\centering            
	\subfloat[]   
	{
		\includegraphics[width=0.4\textwidth]{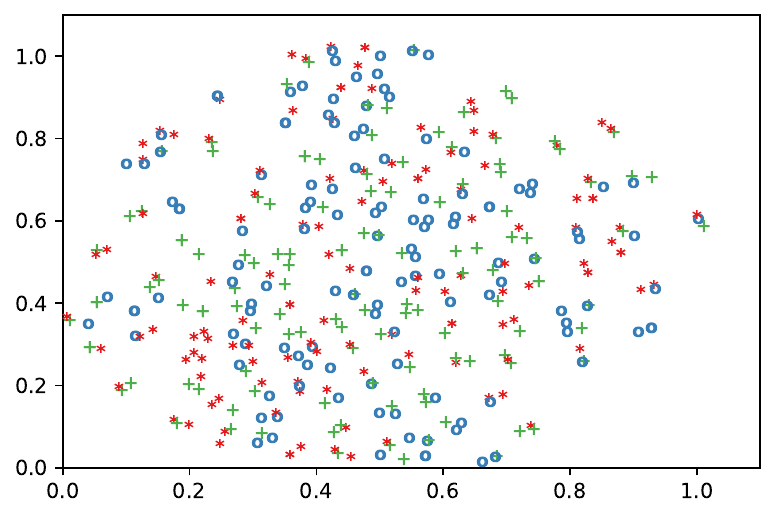}
	} \\
	\subfloat[]
	{
		\includegraphics[width=0.4\textwidth]{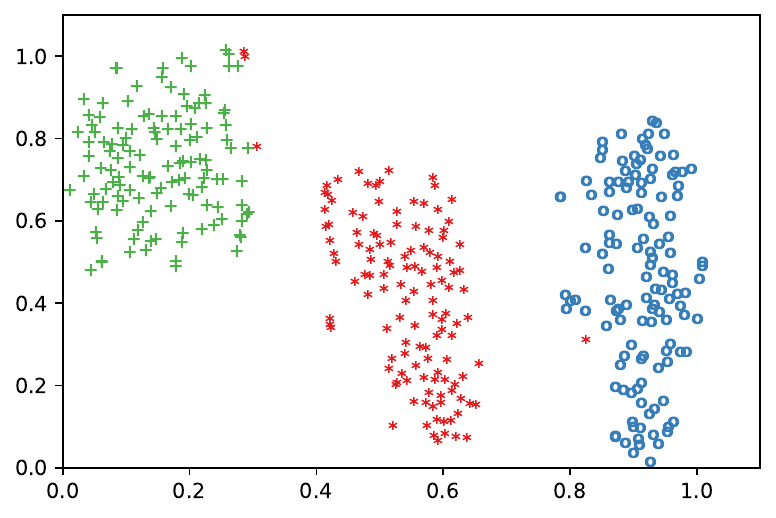}
	}
	\caption{\textcolor{blue}{Visual comparison results of unimodal feature distributions before and after employing MSFE. (a) The fifth-level of features from the Baseline. (b) The fifth-level of features from our proposed MSN.}}    
	\label{fig_1012}            
\end{figure}

\subsubsection{Qualitative results for our proposed MSN with different types of inputs}
The visualization results shown in Fig. \ref{fig_100} can further demonstrate the plausibility of our proposed AM SOD task.
Fig. \ref{fig_100} illustrates some saliency maps under different situations.
As shown in the first row of Fig. \ref{fig_100}, it can be seen that utilizing complementary information from more modalities can constantly improve the performance for most scenes. 
That is, the saliency maps detected from the inputs with two modalities are better than those from the inputs with one modality (Fig. \ref{fig_100}(d)-(f) vs Fig. \ref{fig_100}(g)-(i)). 
And, utilizing complementary information from all modalities achieves the best detection performance (Fig. \ref{fig_100}(j)).

However, some special cases are also displayed in the 2$^{nd}$, 3$^{rd}$ and 4$^{th}$ rows of Fig. \ref{fig_100}.
Specifically, the RGB image quality is superior in the  2$^{nd}$ row, the thermal image quality is superior in the 3$^{rd}$ row and the depth image quality is superior in the 4$^{th}$ row.
It can be seen that, in such scenarios, the results detected from the unimodal image may have been already be relatively good (i.e., Fig. \ref{fig_100}(d) in the 2$^{nd}$ row, Fig. \ref{fig_100}(e) in the 3$^{rd}$ row and Fig. \ref{fig_100}(f) in the 4$^{th}$ row). 
In such case, utilizing information from more modalities does not always further improve the detection performance (e.g., Fig. \ref{fig_100}(d) vs Fig. \ref{fig_100}(j) in the 2$^{nd}$ row). 
This further indicates that it is not necessary to use all modalities for detection in all scenarios, even if we have information from all modalities, thus verifying the importance and necessity of our proposed AM SOD task.

\subsubsection{Effectiveness of MSFE}

\textcolor{blue}{As shown in Table. \ref{tab_998}, we  verify the effectiveness of our proposed MSFE.  `$w/o$ MID' does not employ the modality indicator. `$w$ MID + $w/o$ MIWG' directly generates modality indicator weights from the input modality indicators without using MIWG. '$w$ MID + $w$ MIWG' denotes the model that further employs MIWG on top of `$w$ MID + $w/o$ MIWG'. `$w$ MID + $w$ MIWG + $w$ $L_{modality}$' denotes the final version of MSFE by further employing the cross-modality loss function $L_{modality}$.}

The resuts of `$w$ MID + $w/o$ MIWG' indicate that directly generating feature selection weights cannot effectively select discriminative unimodal features according to the input modalities, since the input modality indicators are unrelated to  those extracted features. Furthermore, by virtue of MIWG, `$w$ MID + $w/o$ MIWG' can explore the interactions between the modality indicators with those extracted features, thus abling to select discriminative unimodal features according to the input modalities. Finally, our proposed MSFE can generate more modality-distinguishable weights for unimodal feature selection by using the cross-modality loss function $L_{modality}$.

\begin{table*}[!t]
	\renewcommand{\arraystretch}{1.3}
	\caption{\textcolor{blue}{Quantitative results from different variants of our proposed MSFE.}} 
	\label{tab_998}
	\centering
	\resizebox{\textwidth}{!}{
		\begin{tabular}{c|cc|cc|cc|cc|cc|cc|cc|cc}
			\hline 
			\multirow{4}{*}{Models}	 & \multicolumn{14}{c|}{Sole} &   \multicolumn{2}{c}{Joint} \\		
			\cline{2-17} 
			&\multicolumn{2}{c|}{RGB} & \multicolumn{2}{c|}{D}  & \multicolumn{2}{c|}{T} & \multicolumn{2}{c|}{RGB-D} & \multicolumn{2}{c|}{RGB-T}& \multicolumn{2}{c|}{D-T}& \multicolumn{2}{c|}{RGB-D-T} & \multicolumn{2}{ c}{ALL} \\
			\cline{2-17} 
			&$\mathcal{M}\downarrow$ &$F_\beta\uparrow$ &$\mathcal{M}\downarrow$ &$F_\beta\uparrow$ &$\mathcal{M}\downarrow$ &$F_\beta\uparrow$ &$\mathcal{M}\downarrow$ &$F_\beta\uparrow$ &$\mathcal{M}\downarrow$ &$F_\beta\uparrow$ &$\mathcal{M}\downarrow$ &$F_\beta\uparrow$ &$\mathcal{M}\downarrow$ &$F_\beta\uparrow$&$\mathcal{M}\downarrow$ &$F_\beta\uparrow$ \\
			\hline 
			$w/o$ MID 
			&0.059&0.779&0.097&0.640&0.065&0.711&0.045&0.825&0.043&0.807&0.0084&0.683&0.0051&0.787&0.055&0.799	  
			\\
			$w$ MID + $w/o$ MIWG  			  &0.059&0.771&0.085&0.654&0.049&0.752&0.048&0.832&0.041&0.817&0.0075&0.721&0.0041&0.821&0.055&0.805
			\\
			$w$ MID + $w$ MIWG    &0.055&0.799&0.065&0.696&0.047&0.764&0.035&{0.850}&0.039&0.831&0.0068&0.736&0.0040&0.837&0.051&0.816
			\\  
			$w$ MID + $w$ MIWG + $w$ $L_{modality}$ &\textbf{0.055}  &\textbf{0.803} &\textbf{0.064} &\textbf{0.700} & \textbf{0.044} & \textbf{0.769}   & \textbf {0.035} & \textbf{0.850} &\textbf{0.038} &\textbf{0.833} &\textbf{0.0065} & \textbf{0.740} &\textbf{0.0035} &  \textbf{0.841}  &\textbf{0.049}     &\textbf{0.816}  \\  
			
			\hline
	\end{tabular}}
\end{table*}

\begin{table*}[!t]
	\renewcommand{\arraystretch}{1.3}
	\caption{\textcolor{blue}{Quantitative results of different variants of DFM.}
		} 
	\label{tab_999}
	\centering
	\resizebox{\textwidth}{!}{
		\begin{tabular}{c|cc|cc|cc|cc|cc|cc|cc|cc}
			\hline 
			\multirow{4}{*}{Models}	 & \multicolumn{14}{c|}{Sole} &   \multicolumn{2}{c}{Joint} \\		
			\cline{2-17} 
			&\multicolumn{2}{c|}{RGB} & \multicolumn{2}{c|}{D}  & \multicolumn{2}{c|}{T} & \multicolumn{2}{c|}{RGB-D} & \multicolumn{2}{c|}{RGB-T}& \multicolumn{2}{c|}{D-T}& \multicolumn{2}{c|}{RGB-D-T} & \multicolumn{2}{ c}{ALL} \\
			\cline{2-17} 
			&$\mathcal{M}\downarrow$ &$F_\beta\uparrow$ &$\mathcal{M}\downarrow$ &$F_\beta\uparrow$ &$\mathcal{M}\downarrow$ &$F_\beta\uparrow$ &$\mathcal{M}\downarrow$ &$F_\beta\uparrow$ &$\mathcal{M}\downarrow$ &$F_\beta\uparrow$ &$\mathcal{M}\downarrow$ &$F_\beta\uparrow$ &$\mathcal{M}\downarrow$ &$F_\beta\uparrow$&$\mathcal{M}\downarrow$ &$F_\beta\uparrow$ \\
			\hline 
			$w/o$ DFM &0.064&0.756&0.094&0.634&0.082&0.673&0.044&0.808&0.045&0.797&0.0096&0.663&0.0057&0.763&0.057&0.778   \\
			$w$ DFM-Global  			  &0.056&0.779&0.073&0.683&0.065&0.711&0.041&0.825&\textbf{0.035}&0.817&0.0084&0.683&0.0051&0.787&0.050&0.799
			\\
			$w$ DFM-Channel  &\textbf{0.055}  &\textbf{0.803} &\textbf{0.064} &\textbf{0.700} & \textbf{0.044} & \textbf{0.769}   &  \textbf{0.035} &\textbf{0.850} &0.038 &\textbf{0.833} &\textbf{0.0065} & \textbf{0.740} &\textbf{0.0035} &  \textbf{0.841}  &\textbf{0.049}     &\textbf{0.816}  \\  
			\hline
	\end{tabular}}
\end{table*}

\subsubsection{Effectiveness of dynamic fusion}

\textcolor{blue}{As shown in Table. \ref{tab_999}, we further verify the effectiveness of our proposed DFM. '$w/o$ DFM' represents the model where the proposed DFM is replaced with a mean operation. '$w$ DFM-Global' stands for the model that directly takes the features of each modality as a token. '$w$ DFM-Global' explores the global interactions among modalities without considering their channel-wise interactions. '$w$ DFM-Channel' denotes our final module.}
	
\textcolor{blue}{It can be observed that, compared to '$w/o$ DFM' (mean operation), '$w$ DFM-Global' obtains better results by exploring global relations across modalities. Furthermore, '$w$ DFM-Channel' achieves the best performance by further exploiting channel-wise interactions across modalities.}

\subsubsection{Visualization of modality indicator weights}

To further prove the effectiveness of our proposed modality switch feature extractor (MSFE), as shown in Fig. \ref{fig_1012}, we visualize the distributions of the fifth level of unimodal features extracted from 128 RGB images, depth images and thermal images, respectively, by the `Baseline' and our proposed MSFE. As shown in Fig. \ref{fig_1012}(a), those extracted features of different modalities are mixed together if without using our proposed MSFE. While, as shown in Fig. \ref{fig_1012}(b), those extracted features of different modalities are well separated by using our proposed MSFE. This indicates that our proposed MSFE can adaptively extract unimodal features according to their modalities.

\section{Future works}

\textcolor{blue}{ For the AM SOD task, we propose a preliminary solution, \emph{i.e.} MSN, which obtains some good results, but still faces many issues. Especially, our proposed model cannot achieve the best performance in many settings. This may be due to the following reasons. First, our feature extractor, MSFE, is based on CNN network, which extracts unimodal features from different modalities in a feature selection way. This implicitly narrows the feature space for representing different modalities. Actually, the idea of using modality prompts on top of Transformer in VScode \cite{vscode} may be an better way for addressing such issues. Secondly, our training strategy is also sub-optimal, which does not allow the interactions among one-modality inputs, two-modality inputs and three-modality inputs.  We also eagerly anticipate that a greater number of scholars will embark on exploring the topic and contribute to the advancement of this field. Accordingly, we will provide access to the code of our proposed MSN and our AM-XD datasets after the article is published. Please follow our github web page at 'https://github.com/nexiakele/AMSODFirst'. Such information has also been added to our revision. }

\section{Conclusion}

In this paper, we present an AM SOD, a new SOD task that aims to identify salient objects by taking images of arbitrary numbers and modalities from the same scene as inputs. 
Particularly, we develop the first AM SOD model named modality switch network (MSN) on top of the analysis of the major challenges faced by AM SOD, and is achieved by our well-designed  modality switch feature extractor (MSFE) and dynamic fusion module (DFM).
Benefiting from the proposed MSFE associated with the corresponding modality indicators, those unimodal features from different types of images can be effectively extracted by using one feature extractor.
Meanwhile, owing to the proposed DFM, the complementary information within the features across the input images of arbitrary numbers and modalities can be dynamically captured for final saliency prediction.
With the collaboration of the proposed components mentioned above, our proposed method breaks the limitations imposed by the modality types and modality numbers on the saliency algorithm.
This reduces the additional cost of designing multiple SOD algorithms with different types of inputs for devices equipped with multiple cameras, meeting the demands of future scenario applications.
Apart from the proposed MSN, we have also developed a new dataset named AM-XD.
Overall, these provide a baseline model as well as a training and testing platform for supporting more research in the future.

\section*{Acknowledgments}

This work is supported by the China Postdoctoral Science Foundation under Grant No.2023M742745 and the Postdoctoral Fellowship Program of CPSF under Grant No.GZB20230559.
It is also supported by the National Natural Science Foundation of China under Grant No.61773301, the Shaanxi Innovation Team Project under Grant No.2018TD-012 and the State Key Laboratory of Reliability and Intelligence of Electrical Equipment No.EERI KF2022005, Hebei University of Technology.


\bibliographystyle{IEEEtran}
\bibliography{IEEEabrv,tex}

\begin{IEEEbiography}[{\includegraphics[width=1in,height=1.25in,clip,keepaspectratio]{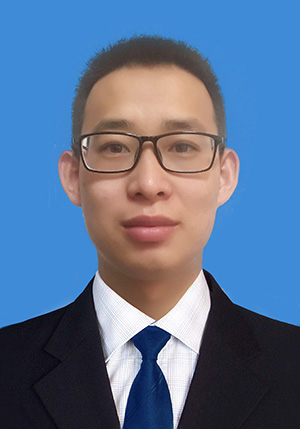}}]{Nianchang Huang}
received the B. S. degree and the M. S. degree from Qingdao University of Science and Technology, Qingdao, China, in 2015 and 2018, and the Ph.D. degree in School of Mechano-Electronic Engineering, Xidian University, China, in 2022. He is currently a lecturer with the Automatic Control Department, Xidian University, China. His research interests include deep learning and multi-modal image processing in computer vision.
\end{IEEEbiography}


\begin{IEEEbiography}[{\includegraphics[width=1in,height=1.25in,clip,keepaspectratio]{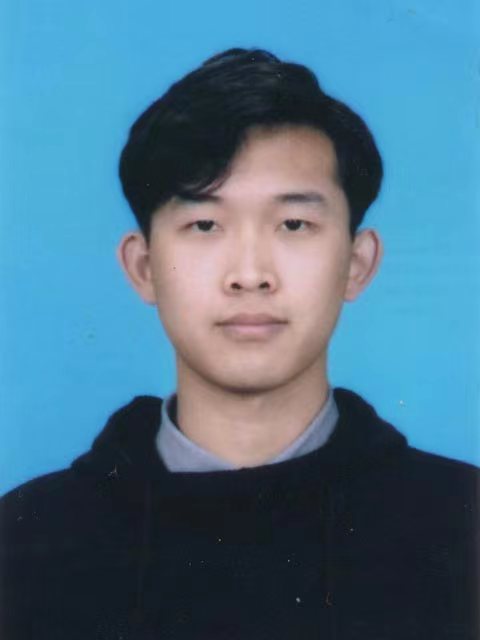}}]{Yang Yang}
	received his B. S. degree from Chang’an University, Xi’an, China, in 2019. He is currently pursuing the Ph.D. degree in School of Mechano- Electronic Engineering, Xidian University, China. His current research interests include multi-modal image processing and deep learning.
\end{IEEEbiography}


\begin{IEEEbiography}[{\includegraphics[width=1in,height=1.25in,clip,keepaspectratio]{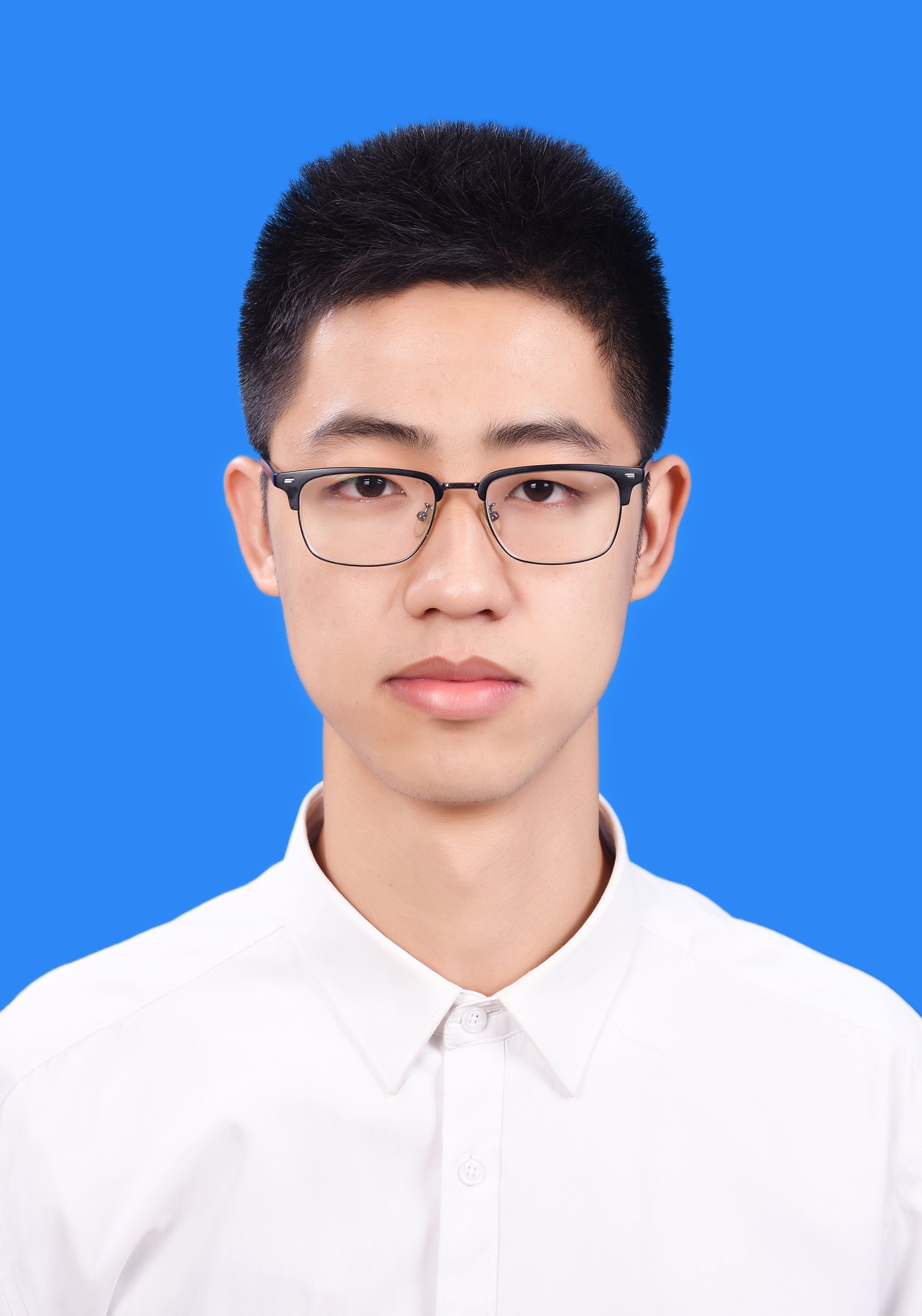}}]
	{Ruida Xi} received the B.S. degree from Xi'an University of Technology, Shaanxi, China, in 2021. He is currently pursuing the Ph.D. degree in School of Mechano-Electronic Engineering, Xidian University, China. His research interests include multi-modal image processing and deep learning.
\end{IEEEbiography}

\begin{IEEEbiography}[{\includegraphics[width=1in,height=1.25in,clip,keepaspectratio]{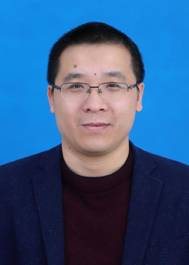}}]{Qiang Zhang}
	received the B.S. degree in automatic control, the M.S. degree in pattern recognition and intelligent systems, and the Ph.D. degree in circuit and system from Xidian University, China, in 2001,2004, and 2008, respectively. He was a Visiting Scholar with the Center for Intelligent Machines, McGill University, Canada. His current research interests include image processing, pattern recognition.
\end{IEEEbiography}

\begin{IEEEbiography}[{\includegraphics[width=1in,height=1.25in,clip,keepaspectratio]{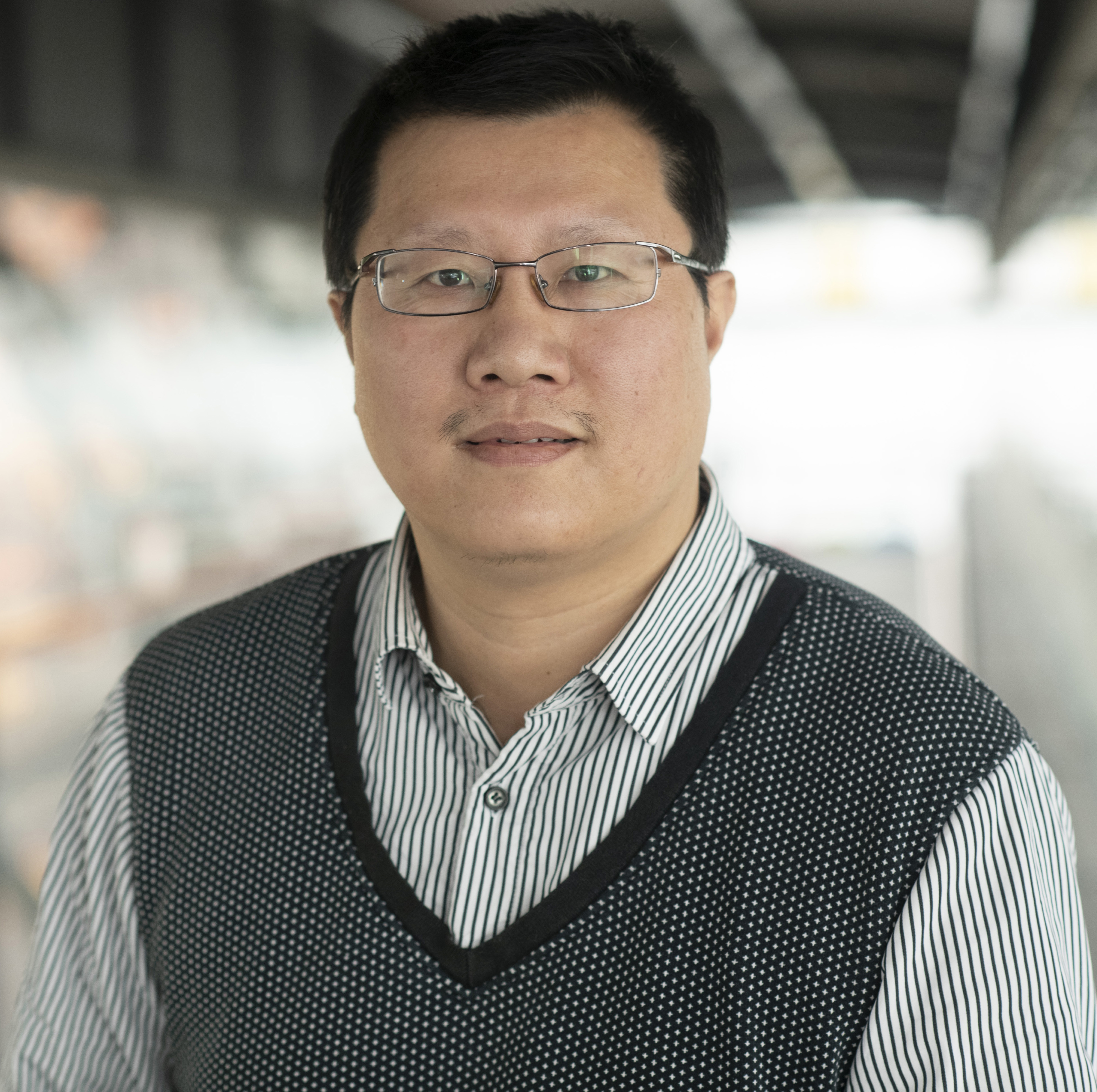}}]{Jungong Han}
	is Chair Professor in Computer Vision at the Department of Computer Science, the University of Sheffield, UK. He also holds an Honorary Professorship at the University of Warwick, UK. Previously, he was Chair Professor and Director of Research of the Computer Science department with Aberystwyth University, UK; Data Science Associate Professor with the University of Warwick; and Senior Lecturer in Computer Science with Lancaster University, UK. His research interests span the fields of video analysis, computer vision, and applied machine learning.
\end{IEEEbiography}

\begin{IEEEbiography}[{\includegraphics[width=1in,height=1.25in,clip,keepaspectratio]{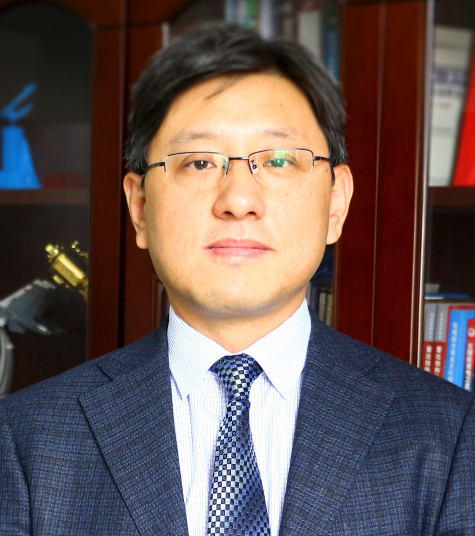}}]{Jin Huang}
	received the Ph.D. degree in mechanical engineering from Xidian University, Xi'an, China, in 1999. He worked with the Department of Mechanical Engineering, University of British Columbia, Vancouver, BC, Canada, as a Visiting Researcher in 2001-2002. He is a Professor, the Dean of the School of Electro-Mechanical Engineering, Xidian University, and the Director with the Key Laboratory of Electronic Equipment Design, Minister of Education. He also is the Fellow of the Chinese Institute of Electronics and served as the Deputy Secretary General of the Electro-mechanical Engineer Society of China. He has authored/co-authored more than 100 papers in various peer reviewed journals and conference proceedings, and holds more than 50 patents. His research interests include  exible electronics, mechatronics, and 3-D printing.
\end{IEEEbiography}
\end{document}